\documentclass[acmtog]{acmart}
\acmSubmissionID{}

\usepackage{booktabs} 
\usepackage{multicol}
\usepackage{multirow}
\usepackage{physics}

\usepackage[ruled]{algorithm2e} 

\SetAlFnt{\small}
\SetAlCapFnt{\small}
\SetAlCapNameFnt{\small}
\SetAlCapHSkip{0pt}

\makeatletter
\DeclareRobustCommand\onedot{\futurelet\@let@token\@onedot}
\def\@onedot{\ifx\@let@token.\else.\null\fi\xspace}

\def\eg{\emph{e.g}\onedot}

\makeatother

\acmJournal{TOG}

\def\Pnorm{\ensuremath{P_N}}
\def\Pnormplus{\ensuremath{P_N^+}}
\def\W{\ensuremath{{W}}}
\def\Wplus{\ensuremath{{W}+}}
\def\Z{\ensuremath{Z}}

\def\jcf{\color{black}}

\renewcommand\footnotetextcopyrightpermission[1]{}
\begin{document}
\title{Improved StyleGAN Embedding: Where are the Good Latents?}

\begin{abstract}
StyleGAN is able to produce photorealistic images that are almost indistinguishable from real photos.
The reverse problem of finding an embedding for a given image poses a challenge.
Embeddings that reconstruct an image well are not always robust to editing operations. 
In this paper, we address the problem of finding an embedding that both reconstructs images and also supports image editing tasks.  First, we introduce a new normalized space to analyze the diversity and the quality of the reconstructed latent codes. This space can help answer the question of where good latent codes are located in latent space. 
Second, we propose an improved embedding algorithm using a novel regularization method based on our analysis. 
Finally, we analyze the quality of different embedding algorithms.
We compare our results with the current state-of-the-art methods and achieve a better trade-off between reconstruction quality and editing quality.
\end{abstract}

%
\author{Peihao Zhu}
\affiliation{%
	\institution{KAUST}
	\country{Saudi Arabia}
}
\email{peihao.zhu@kaust.edu.sa}

\author{Rameen Abdal}
\affiliation{%
	\institution{KAUST}
	\country{Saudi Arabia}
}
\email{rameen.abdal@kaust.edu.sa}

\author{Yipeng Qin}
\affiliation{%
	\institution{Cardiff University}
	\country{UK}
}
\email{qiny16@cardiff.ac.uk}

\author{John Femiani}
\affiliation{%
	\institution{Miami University}
	\streetaddress{510 E. High St}
	\city{Oxford}
	\state{OH}
	\postcode{45056}
	\country{USA}
}
\email{femianjc@miamioh.edu}

\author{Peter Wonka}
\affiliation{%
	\institution{KAUST}
	\country{Saudi Arabia}
}
\email{pwonka@gmail.com}
	
%

%
%


\maketitle
\thispagestyle{empty}
\pagestyle{plain} 
\section{Introduction}



GAN inversion (embedding) refers to the task of computing a latent code for a given input image. 
The goal of the embedding is typically to perform some subsequent image processing tasks, such as image interpolation or semantic image editing.
Due to the high visual quality of generated images and comparatively lightweight architecture, most recent papers build on StyleGAN~\cite{stylegan2019} and StyleGAN2~\cite{styleganv2-2020,Karras2020ada} to develop GAN inversion algorithms.
The resulting embedding methods represent a trade-off of two main concerns: the reconstruction quality and the editing quality. 
The reconstruction quality considers how similar an embedded image is to the input image.
The editing quality describes the visual quality {\jcf and realism} of images after performing editing operations in latent space, e.g., editing the pose, lighting, or age of in face images~\cite{harkonen2020ganspace}. 

\begin{figure}[th]
    \centering
    \includegraphics[width=\linewidth]{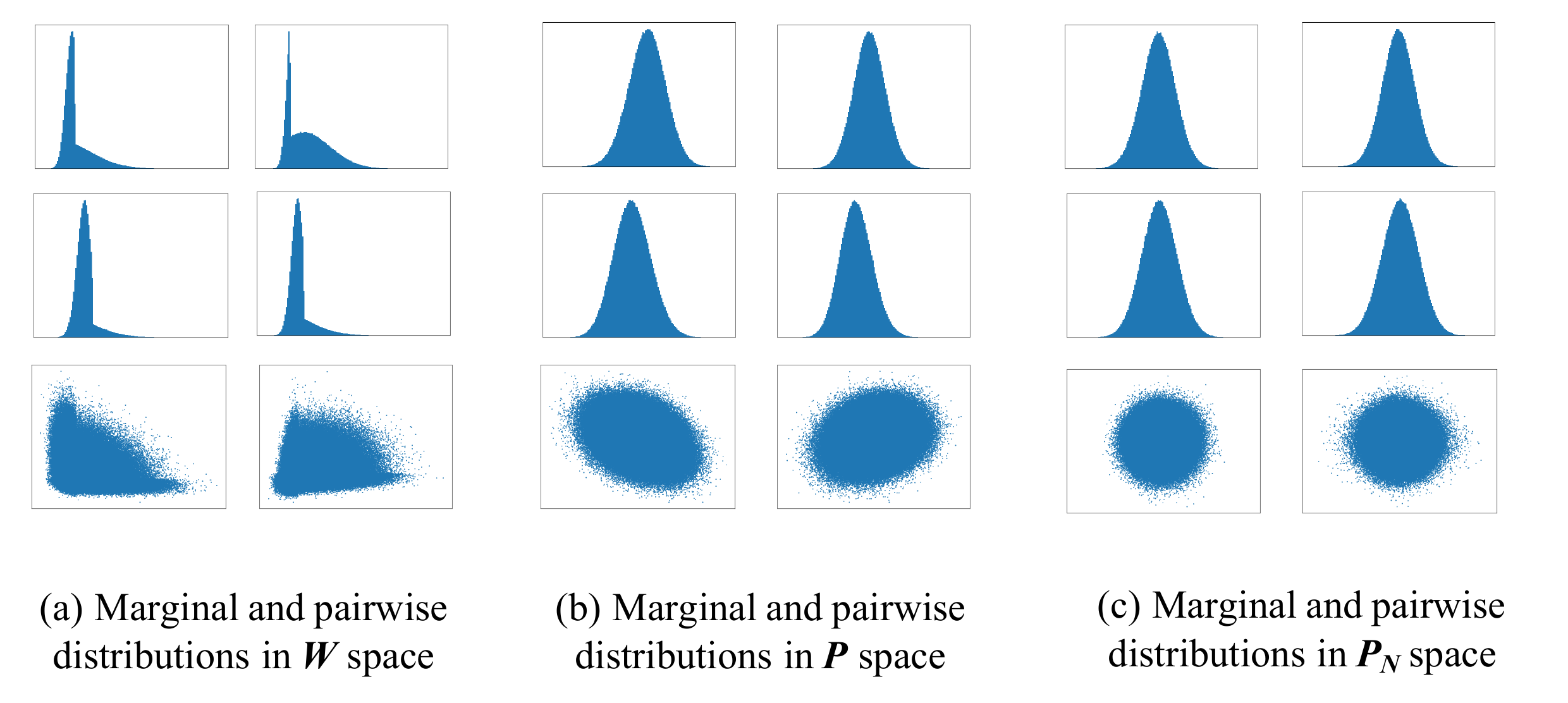}
    \caption{Marginal and pair-wise distributions of different spaces in StyleGAN. First and second rows: the marginal distribution of a randomly picked latent variable. Third row: the pairwise joint distribution of two randomly picked latent variables.}
    \label{fig:distribution}
\end{figure}

In this paper, we set out to provide some analysis of existing embedding methods {\jcf and use that analysis as the basis for an improved image-to-StyleGAN embedding method}. 
One important consideration is the choice of the embedding space, e.g., \Z{} space, \W{} space, or \Wplus{} space. 
{\jcf These spaces each impose trade-offs -- the \Z{}~space, used when training StyleGAN or StyleGAN2, restricts latent codes to a hypersphere. The \W{}~space does not have a sphere constraint and uses a mapping network to disentangle latent codes. The \W{} space is therefore less entangled than the \Z{} space. However, both spaces use 512 dimensional latent codes. This does not provide enough expressiveness to represent all real-world faces well. In \cite{abdal2020image2stylegan++} it was shown that \Wplus{} codes, in which the degrees of freedom are 18 times larger than \W{}~space, are capable of reconstructing images well. However, these additional degrees of freedom make GAN inversion an ill-conditioned problem. Thus, non regularized inversion of \Wplus{} codes can find embeddings in low-density portions of the latent space, so that even if a particular image is reconstructed well, the reconstruction is in a region that will not normally be reached by the mapping network. Empirically, this leads to undesirable images after GAN-based image manipulations such as interpolation or semantic editing. }

 We propose a suitable space, called \emph{\Pnorm{}} space, to perform the analysis. 
In \Pnorm{} space, the distribution of StyleGAN latent codes has a surprisingly simple structure{\jcf, illustrated in Fig.~\ref{fig:distribution}, which we approximate by a multivariate normal distribution}.
This enables an explanation of where good latent codes can be found.
A major insight of the analysis is that the quality of a latent code is closely related to the distance from the center of \Pnorm{} space. 
The $L_2$ norm in \Pnorm{} space is a Mahalanobis distance of latent codes, so that $L_2$ regularization in this space will bias embeddings towards {\jcf more densely sampled regions of the GAN latent space}. 
We find that reconstruction quality favors latent codes that are far from the origin in \Pnorm{} space, but editing quality favors latent codes close to the origin {\jcf in \Pnorm{} space}.
This requires a trade-off between reconstruction quality and editing quality.
Investigating the individual dimensions of latent codes in \Pnorm{} space gives a strong understanding of the trade-offs made by current methods.
%

We propose an improved embedding algorithm by introducing a regularizer that encourages embeddings to stay closer to the origin of the \emph{\Pnorm{}} space{\jcf, which is where the \W{} latent-codes of training data have the highest density. We further extend this to the much higher capacity \Wplus{} space, introducing a corresponding $\Pnormplus{}$ regularizer.} The advantage of our regularizer is that it explicitly controls the distance to the origin and does not have undesirable side effects. {\jcf In particular, a regularizer will bias solutions towards more dense and editable portions of latent space, but it will not prevent optimization from finding solutions that reduce the reconstruction error.  Regularizing will slightly reduce reconstruction quality, but much less-so than a constraint. Further, a regularization hyperparameter allows control over the trade-off between reconstruction quality and editability.  }
Finally, we present the first comprehensive evaluation comparing different state-of-the-art embedding algorithms. The evaluation does not just consider reconstruction quality, but also the effect of a variety of edits (pose, lighting, and age) and the impact of embedding algorithms on conditional embedding tasks (super-resolution, image colorization, and inpainting).
Our main contributions are:
\begin{itemize}
    \item the introduction of \Pnorm{} {\jcf and \Pnormplus{}} space for analyzing and regularizing StyleGAN embeddings.
    \item a new regularizer for StyleGAN embedding that provides the best trade-off between reconstruction and editing quality.
    \item a comprehensive evaluation of many state-of-the-art embedding algorithms. Our evaluation not only emphasizes reconstruction, but also the impact on downstream editing tasks.
\end{itemize}

\section{Related Work}

The seminal paper by Goodfellow et al.~\cite{goodfellow2014generative} started an avalanche of GAN papers that together lead to impressive improvements over the last years~\cite{radford2015unsupervised,arjovsky2017wasserstein,karras2017progressive,miyato2018spectral,stylegan2019,styleganv2-2020,Karras2020ada}. Similar to most other work in GAN image embedding, we build on StyleGAN~\cite{styleganv2-2020, stylegan2019, Karras2020ada} due to its exceptional visual quality and comparatively lightweight architecture.

Since the recent inception of Image2StyleGAN (I2S)~\cite{image2stylegan2019} which first proposed a feasible method to embed a given image into the \Wplus{} latent space of the StyleGAN generator, there have been many works trying to improve upon the initial idea. 

Most of the existing methods have the same goal: finding a balance between the reconstruction quality and the editing quality.
For example, I2S++~\cite{abdal2020image2stylegan++} improved the reconstruction quality of I2S by incorporating a noise optimization step to restore the high-frequency details in the input image.
Similarly, StyleGAN2~\cite{styleganv2-2020} uses an additive ramped-down noise to help the latent space exploration. 
These two methods lead to different trade-offs:  I2S++ embeds images into the \Wplus{} space that sacrifices a bit of editing quality to achieve better reconstruction quality; while on the contrary, the StyleGAN2 embeds images in the \W{} space that enables better editing but at the cost of worse reconstruction.
Following a different approach, PIE~\cite{tewari2020pie} first embeds an image into the \W{} space for better editing quality and then improves the reconstruction quality by optimizing the latent obtained in the \Wplus{} space.
Such a two-stage encoding process is also employed by several concurrent works based on encoder networks~\cite{bau2019seeing,zhu2020indomain,guan2020collaborative,bau2020semantic}, which to our knowledge first appears in iGAN~\cite{zhu2016generative}. 
In their methods, an initial latent code is first obtained by passing the input image through a pre-trained StyleGAN encoder. Then, the initial latent code is further optimized to improve its reconstruction quality. Using only the encoder network by itself leads to poor reconstruction quality.

Unlike previous methods, PULSE~\cite{menon2020pulse} formulated embedding as traversing the manifold consisting of good latents. However, they {\jcf constrained} the latents to be on the surface of a hypersphere that only contains a subset of good latents. As a result, their method usually leads to poorer reconstruction quality.
Inspired by PULSE, in this work, we provide a thorough analysis of where the good latents are and how to evaluate an embedding. Based on our analysis, we propose an improved embedding algorithm that outperforms all existing methods.



\begin{figure*}[th]
    \centering
    \includegraphics[width=\linewidth]{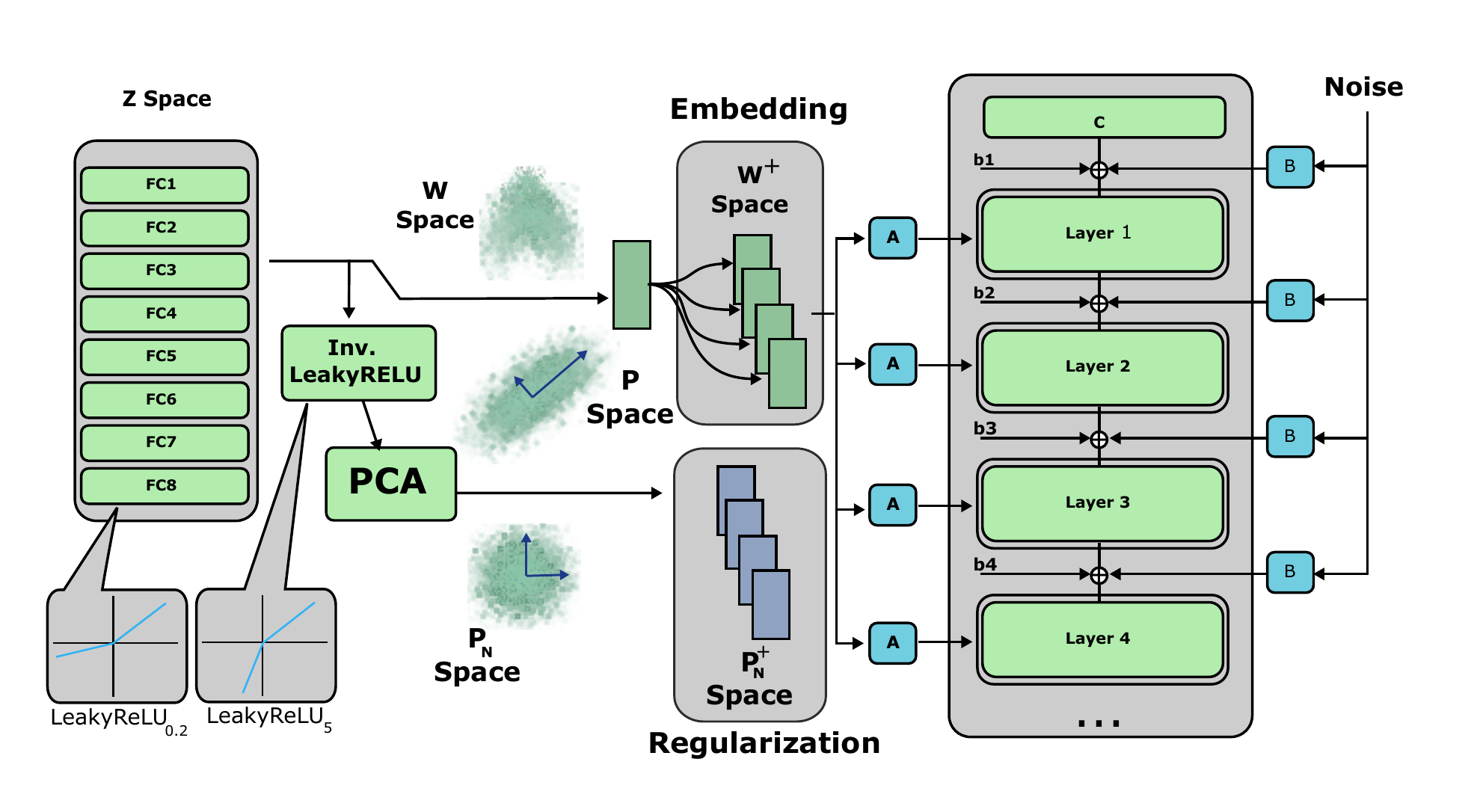}
    \caption{We show six different spaces into which projection of a given image is possible. These are: \Z{} space, \W{} space, \Wplus{} space, $P$ Space, \Pnorm{} space and \Pnormplus{} space. Notice that translation from \W{} to $P$ or $P$ to \W{} space goes through an invertible function (Leaky ReLU). }
    \label{fig:EmbeddingSpaces}
\end{figure*}

\section{Which Space to Use: a Statistical Analysis of Latent Distributions}




The goal of this section is to identify a space where the distribution of latent codes has a simple structure. A suitable space will make it easier to reason about good and bad latent codes for the StyleGAN generator~\cite{stylegan2019,styleganv2-2020}.
%
%
To find such a latent space, we conduct a statistical analysis on the latent distributions in different latent spaces {\jcf illustrated in } Fig.~\ref{fig:EmbeddingSpaces}, {\jcf including \W{}, \Pnorm{}, \Wplus{}, and \Pnormplus{} spaces}.

\paragraph{$\emph{W}$ Space:}
The \W{} space proposed in StyleGAN~\cite{stylegan2019} is the most straightforward choice of latent space.
Therefore, we start our investigation by 
{\jcf analyzing the distribution of} $1$ a million latent codes in \W{} space and visually checking the statistics of their marginal distributions.
{\jcf We sample this space by first generating random samples in \Z{}~space, and then forward-propagating through the mapping layers of StyleGAN2. We use the same input distribution as StyleGAN, in which the latent-codes in \Z{}~space are projected onto a hypersphere before propagating them through the mapping network.}
As Fig.~\ref{fig:distribution} (a) shows, the marginal distributions are heavily skewed.

\begin{figure}[htbp]
\includegraphics[width=0.49\columnwidth]{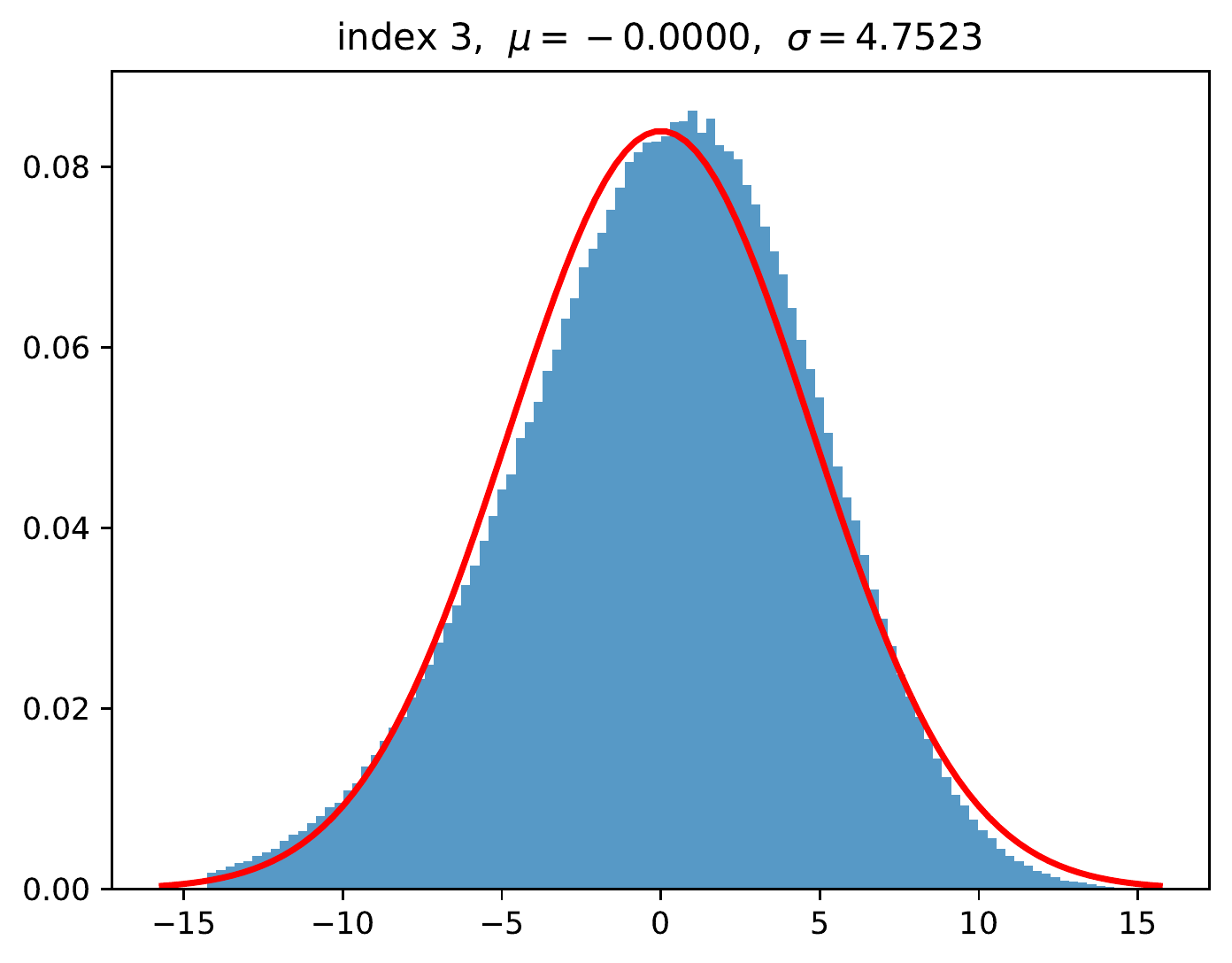}
\includegraphics[width=0.49\columnwidth]{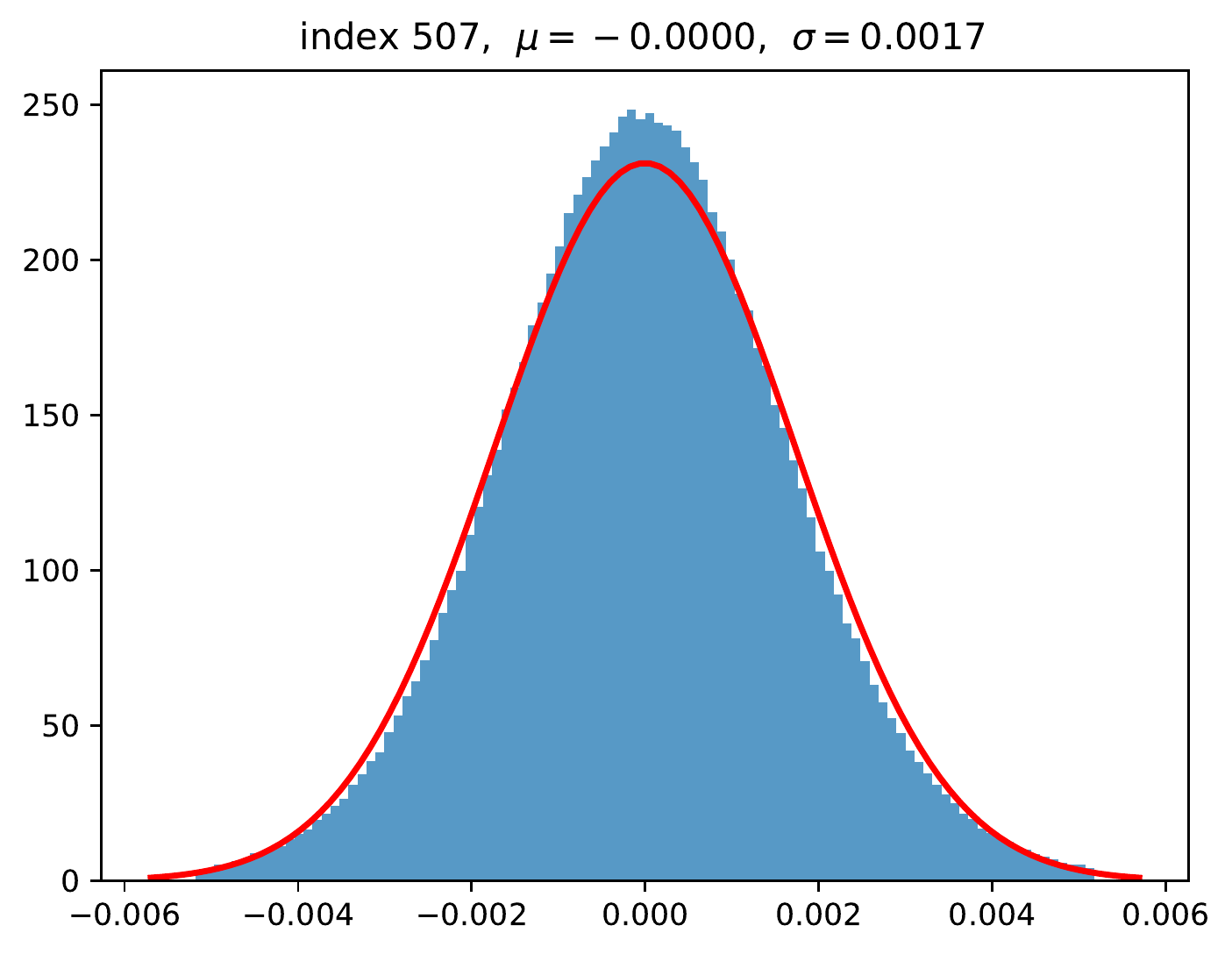}
\caption{Examples of marginal distributions of the latent-codes, with a best-approximating normal distribution indicated in red. The distribution is clearly not normal based on skew (principle axis 3, left subplot) and kurtosis (principle axis 570, right subplot), however even the marginals that fail a normality test are still visually quite similar to a normal distribution.}
\label{fig:marginals}
\end{figure}

\paragraph{$\emph{P}$ Space:}
To get rid of the skewness of marginal distributions, we transform the \W{} space to the $P$ space by inverting the last Leaky ReLU layer in the StyleGAN mapping network. 
Since the last Leaky ReLU uses a slope of $\frac{1}{5}$ we invert it using  
\begin{equation}
    \mathbf{x} = \mathrm{LeakyReLU_{5.0}}(\mathbf{w})
    \label{eq:p_space}
\end{equation}
where $\mathbf{w}$ and $\mathbf{x}$ are latent codes in \W{} and $P$ space respectively.
Similar to those in \W{} space, we plot the marginal distributions of latent codes in $P$ space and observed that they appear to follow a simple Gaussian-like distribution.
We plot the pairwise joint distribution of latent codes (See Fig.~\ref{fig:distribution}b) and observe that the dimensions are not statistically independent. 
{\jcf 
Based on the marginal plots, we suspected that the distribution may be a Multivariate Normal Distribution, so we performed a Henze-Zirkler~\cite{henze1990class} test for multivariate normality in 512 dimensions, using 1,000 samples, the resulting test indicates that the data is not normal, with a significance of $\alpha=5\%$. We also used a Mardia test~\cite{mardia1974applications,mardia1970measures} of multivariate normality, which is based on a measurement of multivariate skewness,  and kurtosis. The Mardia test had similar results; the distribution is not normal, with a significance of $5\%$. 
We tested the marginal distribution of the data along each principal component axis, and found that 219 of the 512 marginal are normal at $5\%$ significance according to the Mardia test. 
We show two marginal distributions where the Mardia statistic indicates that the data is not normal in Fig~\ref{fig:marginals}.   Since the distribution visually appears to have a unimodal bell-shaped marginal distribution when we look at a single axis, we used a Dip test~\cite{hartigan1985dip} for unimodality and confirmed that the multivariate distribution does indeed have a single peak. Although the distribution is not a multivariate normal distribution, is it unimodal and has correlated features. Visual inspection of Fig~\ref{fig:distribution} and Fig~\ref{fig:marginals} indicates that even if it is not normal, the data could be well-approximated by a normal distribution. Another reason why the data cannot be a true normal distribution is because the mapping network uses an MLP architecture. Such an architecture creates a distribution with compact support, and the generated latent variables cannot be arbitrarily large.
}
This motivates us to introduce the \Pnorm{} Space in which the latent distribution is easier to characterize.

\paragraph{\Pnorm{} Space:}
Our \Pnorm{} space {\jcf is inspired by principal component analysis (PCA) whitening, and} aims to eliminate the dependencies among latent variables in $P$ space.
We define a transformation from $P$ to our \Pnorm{} space as:
\begin{equation}
    \hat{\mathbf{v}} = \mathbf{\Lambda}^{-\frac{1}{2}} \cdot \mathbf{U}^T (\mathbf{x} - \boldsymbol{\mu})
    \vspace*{-2mm}
    \label{eq:p_to_p_norm}
\end{equation}
where $\mathbf{\Lambda}$ is a diagonal matrix of  singular values, $\mathbf{U}$ is an orthogonal matrix of principal component directions, and $\boldsymbol{\mu}$ is a mean vector. 
Intuitively, {\jcf this whitening, or sphering, transform maps} the distribution of each latent variable to be of zero mean and unit variance.
As a result, the latent distribution in our \Pnorm{} space will look like a ball that is isotropic in all directions (Fig.~\ref{fig:distribution}c).

\begin{figure}[htbp]
    \centering
    \includegraphics[width=0.95\columnwidth]{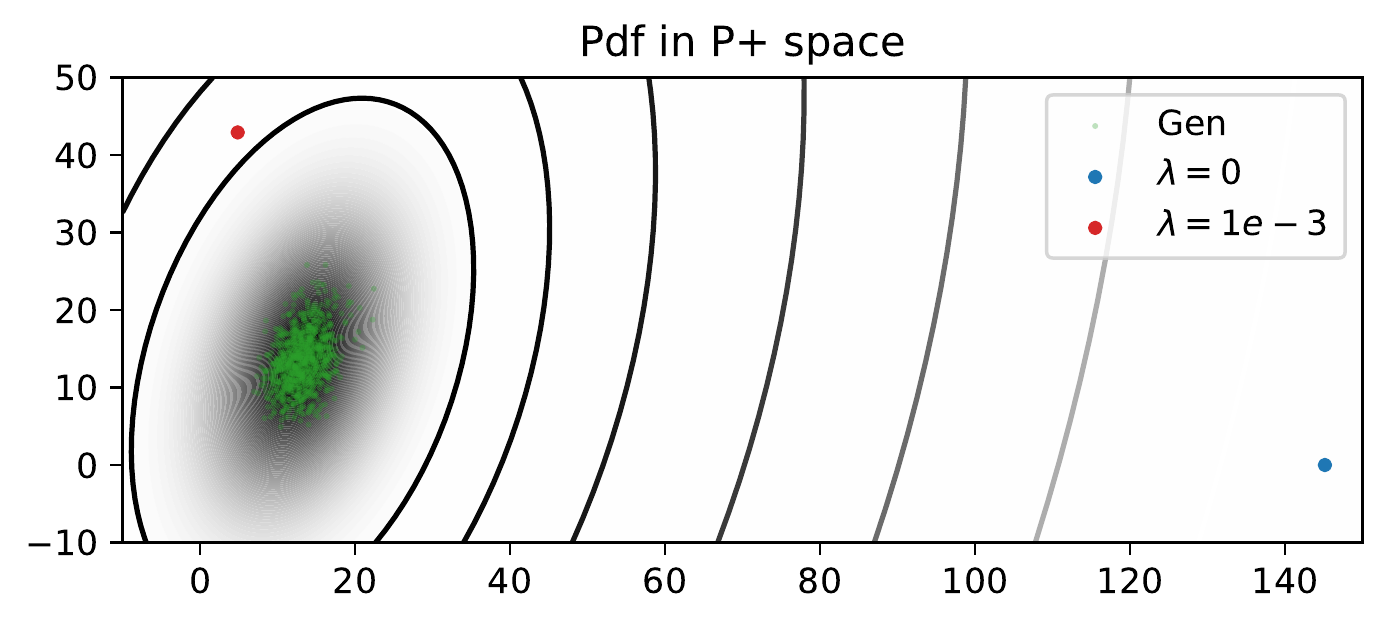}
    \caption{The relationship of latent codes generated by the GAN (green) and two latent codes found by GAN inversion of the same image. The distribution is projected onto the plane containing the two latent-codes and the center of the distribution. We observe that increasing regularization does not shift the latent code straight towards the center, the reconstruction is less regularized along the long axis of the ellipse, and it is closer to the center in the short leg of the ellipse.
    }
    \label{fig:latent-space}
\end{figure}

The $L2$  norm $\|\hat{\mathbf{v}}\|$ is related to the Mahalanobis distance $d_M(\cdot)$ of points $\mathbf{x} \sim N(\boldsymbol{\mu}, \mathbf{\Sigma})$ with the singular value decomposition of symmetric $\mathbf{\Sigma}=\mathbf{U} \mathbf{\Lambda} \mathbf{U}^T$, that is
\begin{align}
    d_M^2(\mathbf{x}) &=  (\mathbf{x}-\boldsymbol{\mu})^T\Sigma^{-1}(\mathbf{x}-\boldsymbol{\mu}) \\
    &= \left(\mathbf{\Lambda}^{-\frac{1}{2}}\mathbf{U}^T(\mathbf{x}-\boldsymbol{\mu}) \right)^2 \\
    & = \hat{\mathbf{v}}^T \hat{\mathbf{v}}
\end{align}
This space has useful properties for regularizing our embedding algorithm because it biases the solution towards the mode $\boldsymbol{\mu}$ of embeddings, and it is more sensitive (regularization has more effect) to directions of low variance {\jcf }in the space used when training the GAN. {\jcf The effect of this space on latent codes is illustrated in Fig~\ref{fig:latent-space}}

\begin{figure}[htbp]
    \centering
    \includegraphics[width=0.95\columnwidth]{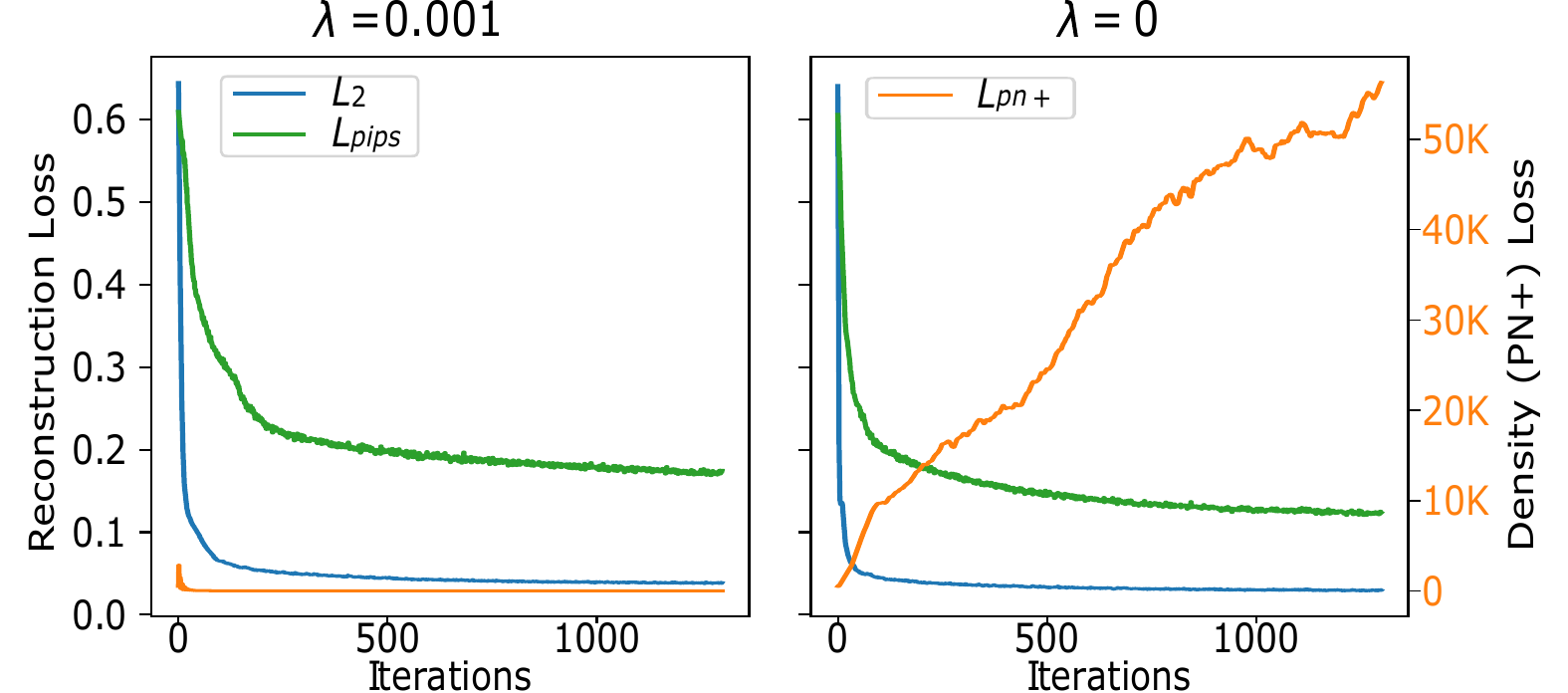}
    \caption{
    The effect of regularization in \Pnormplus{} space vs reconstruction quality. The orange curves are proportional to the negative log-density ($\|\vb{v}\|^2$) of points based on a normal approximation to the distribution of latent codes, the green and blue curves are reconstruction losses.
    Observe that\Pnormplus{} loss can be kept very low with little effect on reconstruction, 
    whereas ignoring the loss causes $\|\vb{v}\|^2$ to be arbitrarily large
    (note that the scale of the $y-$axis on the right is different in order to fit the curve into the plot).
    Shifting the latent codes towards higher-density regions has very little effect on reconstruction quality.}
    \label{fig:effect-of-loss}
\end{figure}

{\jcf We are inspired by generalized Tikhonov regularization, which biases latent codes to more dense portions of the latent space. Unlike truncation, this regularization approach has a much greater effect on subspaces where the input distribution has higher precision, and it has less effect in subspaces where the input distribution has high variance. Simply shifting a latent code towards the origin in \Wplus{} space would, for example, also change the components of the latent code that are important for reconstruction. On the other hand, regularizing based on $d_M$ (i.e. in $P_N$ space) has the desirable effect that it preserves components that also have high variation in the training corpus.

Without regularization, editing tasks are affected because images contain elements that are not modeled by the GAN, so that their embedding can end up in a less-explored region of the embedding space, where the GAN may be more likely to produce low quality results even though the particular image being inverted seems plausible. 
This effect can be observed by comparing the negative log-density (the Mahalanobis distance, $d^2_M=\|\hat{\mathbf{v}}\|^2$) vs the number of optimization steps for GAN inversion with and without regularization, shown in Fig~\ref{fig:effect-of-loss}. Reconstruction quality is improved by a negligible amount, whereas the log-density becomes arbitrarily high. 
An edit operation shifts the embedding to a new location, so if the original embedding was in a high density part of the embedding space the perturbed embedding is also likely to be nearby in a high density region. However, in the low-density portion of the space, a perturbed image will also have a low-density and since it is not the result of inverting a `real world' image it may be perceived as low quality or unrealistic. } 

To verify the validity of our \Pnorm{} space, we sample from a standard Normal distribution in \Pnorm{} space and compare the FID scores we get with those of StyleGAN2~\cite{styleganv2-2020}.  \Pnorm{} space sampling yields an FID of $3.28$ compared to an FID of $2.81$. Such as small difference supports the validity of our \Pnorm{} space. By contrast, fitting a normal distribution to \W{} space and sampling from it yields an FID of $76.63$.
Although good for characterization of latent distributions,
\Pnorm{} space is too restrictive for image embedding. 
To this end, we extend it to \Pnormplus{} space following I2S~\cite{image2stylegan2019}.

\paragraph{\Pnormplus{} Space:}

Similar to the extension from \W{} space to \Wplus{} space in which 18 different 512-dimensional \W{} latents ($\mathbf{w}_i$) are concatenated~\cite{image2stylegan2019},
\begin{equation}
\mathbf{w}^+=\{\mathbf{w}_i\}_{i=1}^{18}
\end{equation} we extend \Pnorm{} space to \Pnormplus{} space as
\begin{equation}
    \mathbf{v}=\{\Lambda^{-1}\mathbf{V}^T (\mathbf{x}_i-\boldsymbol{\mu})\}_{i=1}^{18}
\end{equation}
where $\mathbf{x}_i = \mathrm{LeakyReLU_{5.0}}(\mathbf{w}_i)$.
Each of the latent-codes is used to demodulate the corresponding StyleGAN feature maps at different layers.
We propose to use \Pnormplus{} space to analyze and regularize StyleGAN inversion algorithms.



\section{Improved I2S}
\label{sec:ii2s}





\subsection{Reconstruction Losses}
\label{sec:improved-perceptual-regularizer}
One important aspect of I2S is to use a perceptual regularizer based on VGG-perceptual loss~\cite{simonyan2014very}. However, we find that using LPIPS~\cite{zhang2018perceptual} results in a noticeable improvement.
Following PULSE~\cite{menon2020pulse} we also propose to use bicubic downsampling on the generated images and Lanczos downsampling~\cite{Turkowski_turkowskifilters} on the reference image. We did not use Lanczos for both as it is not differentiable. Nevertheless, we observed that the current setup works better than using bicubic sampling for both cases.

\subsection{\Pnormplus{} Density Based Loss}
Based on our results, it seems most promising to devise a regularizer that penalizes an embedding if it goes too far from the center {\jcf in \Pnormplus{}}. 
Let $I$ be the input image, $L= L_{LPIPS} + L_{2}$ be a loss function consisting of LPIPS and pixel-wise L2 loss terms, $\mathbf{w^+}$ and $\mathbf{v}$ be the latent codes in $W^+$ and \Pnormplus{} space respectively, and $\lambda$ {\jcf is a hyperparameter controlling the amount of regularization}. We have:
    \begin{equation}
        {\mathbf{w}^{+}}^*=\arg \min _{\mathbf{w^+}} L\left(I, G\left(\mathbf{w^+}\right)\right)+\lambda \|\mathbf{v}\|^{2}
        \label{eq:pnorm-plus-L2}
    \end{equation}

\subsection{Implementation Details}
For our method, we used a learning rate of 0.01 with 1300 steps for all images. We used the ADAM optimizer with standard settings to obtain the results.

\begin{figure*}[thpb]
    \centering
    \includegraphics[width=\linewidth]{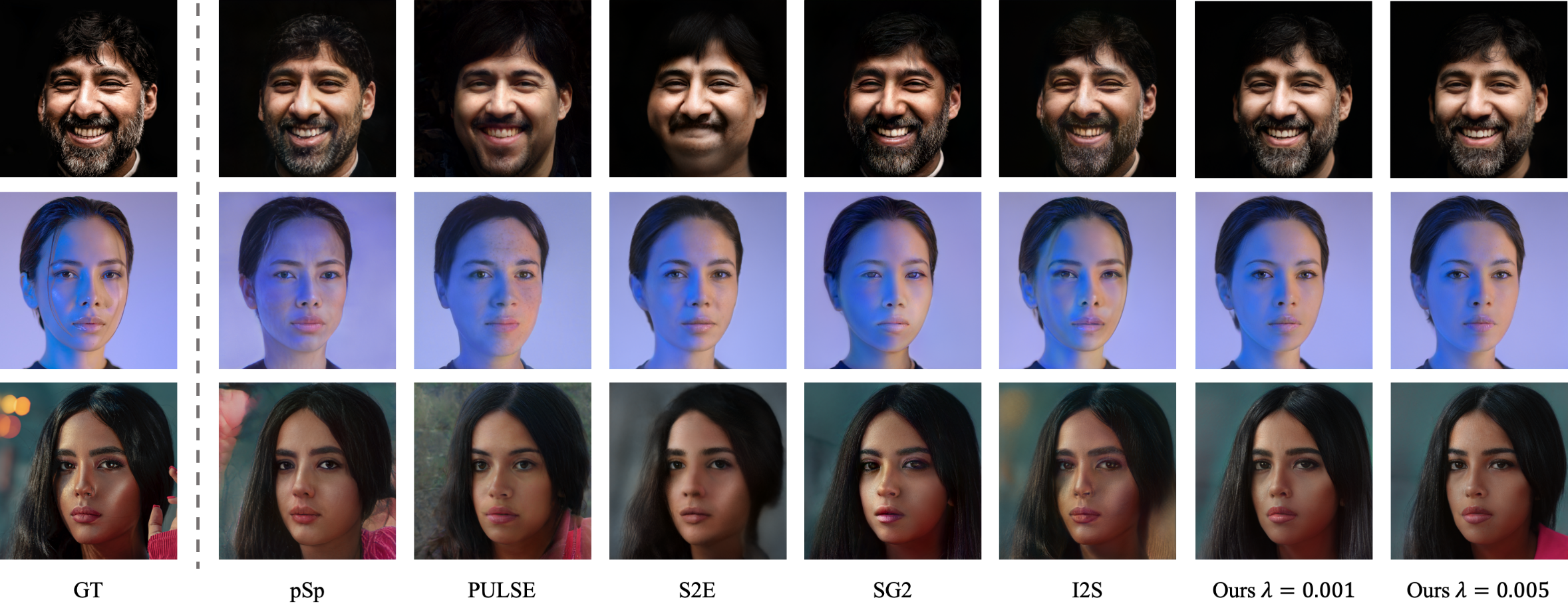}
     \caption{Reconstruction quality of different embedding methods using the StyleGAN2 generator. Close inspection will show that none of the methods can cope with specular reflections. The most accurate reconstruction results are from I2S, which is equivalent to II2S (ours) with $\lambda=0$. However, by increasing $\lambda$ we are able to find a compromise between high reconstruction quality and also realism. For example, the nose-ring on the bottom row is outside the space of the GAN, I2S generates an artifact that II2S does not.}
    \label{fig:recon}
\end{figure*}

   


\section{Evaluation}\label{sec:evaluation}

The purpose of this section is to provide an extensive evaluation and comparison of different embedding algorithms. We propose a sequence of tasks for that purpose.




We extensively tested many methods that have code available and selected the methods that performed best for a more detailed comparison. 
We compare our proposed II2S method (Section~\ref{sec:ii2s}) with the following state-of-the-art methods:
\begin{itemize}
    \item I2S~\cite{image2stylegan2019} is the baseline method we would like to improve upon which embeds images into the \Wplus{} space.
    \item PULSE~\cite{menon2020pulse} 
    proposed embedding onto the surface of a hypersphere in \Z{} space.
    \item StyleGAN2~\cite{styleganv2-2020} embeds images into the W space with the help of noise regularization.
    \item StyleGAN2Encoder~\cite{stylegan2encoder} embeds images into the \Wplus{} space using logcosh image loss, MS-SSIM loss, L1 penalty on latents and several other losses. In addition, they also use early stopping to prevent overfitting.
    \item pSp~\cite{richardson2020encoding} proposes a novel encoder architecture to map the given image to the \Wplus{} space. The method is concurrently developed and only available on arXiv. However, we include it to make the comparison stronger. 
\end{itemize}

Note that we updated I2S and PULSE to use StyleGAN2 and adjusted the hyperparameter settings for I2S to limit overfitting. We do not compare to the IDInvert method~\cite{zhu2020indomain} and Multi-code embedding~\cite{gu2020image} because they only work on the original StyleGAN architecture~\cite{stylegan2019} at a low resolution of $256\times256$. We also do not use noise embedding as proposed by I2S++~\cite{abdal2020image2stylegan++} because it is too hard to control with respect to editing quality.

For the evaluation, {\jcf  it is important to use a set of test images that were not part of the set used to train SyleGAN2.} We collected a dataset of 120 images from the website Unsplash. The remaining subsections describe the tasks and metrics used to evaluate the system.

\subsection{Reconstruction Quality}
\label{sec:reconstruction}

In this work, we measure the reconstruction quality of an embedding both i) absolutely using RMSE, PSNR and ii) perceptually using SSIM, VGG perceptual similarity, LPIPS perceptual similarity and the FID~\cite{heusel2017gans} score between the input and embedded image. The results are shown in Table~\ref{tab:similarity quantitative table}. 
Our method II2S with $\lambda = 0.001$ and I2S are either ranked first or second across all metrics, except SSIM.
However, we note that the metrics are less important than perceptual user evaluations.
Table \ref{tab:user-studies} presents a user study for a stronger regularized setting with $\lambda = 0.005$. 
This is our recommended setting for the best compromise between editing quality and reconstruction quality. 
Here,  II2S reconstruction results were accepted by users more than any competing method except I2S.
In Fig.~\ref{fig:recon}, we show the reconstruction quality.
Notice that other methods can have visual artifacts or differ substantially from the original for some examples when viewing the embedding results at higher resolution.

\begin{table}[thpb]

                       



\centering
\footnotesize
\begin{tabular}{cccccccc} \toprule

\multirow{1}{*}{Method}  & \multicolumn{1}{c}{SSIM}  & \multicolumn{1}{c}{RMSE}  & \multicolumn{1}{c}{PSNR} & \multicolumn{1}{c}{VGG} & \multicolumn{1}{c}{LPIPS}  & \multicolumn{1}{c}{FID}  \\ \midrule
                       
pSp~\cite{richardson2020encoding} & 0.81 &0.09 &21.35 &0.91 &0.30 &54.80\\ 

PUL~\cite{menon2020pulse}  &0.81	&0.09	&21.63	&1.05	&0.37    &64.63 \\

S2E~\cite{stylegan2encoder} &\underline{\textbf{0.84}}	&0.08	&22.50	&0.93	&0.32 &70.70\\

SG2~\cite{styleganv2-2020} &0.79	&0.12	&19.47	&0.80	&\underline{0.21} &46.49\\

I2S~\cite{image2stylegan2019} &\underline{0.83}	&\underline{\textbf{0.07}}	&\underline{\textbf{23.45}}	&\underline{\textbf{0.63}}	&\underline{0.21}   	&\underline{\textbf{36.60}}    \\ \midrule

$\lambda=0.01$	&0.81	&0.09	&20.64	&0.90	&0.27  &53.89\\

$\lambda=0.005$ &0.82	&0.08	&21.45	&0.85	&0.25 &49.91\\

$\lambda=0.001$ &\underline{0.83}	&\underline{\textbf{0.07}}	&\underline{23.00}	&\underline{0.76}	&\underline{\textbf{0.20}}	&\underline{43.99}\\

\bottomrule
\end{tabular}
\caption{
Quantitative comparison of reconstruction quality using similarity metrics SSIM, RMSE, PSNR, VGG~\cite{simonyan2014very}, LPIPS~\cite{zhang2018perceptual}, and FID~\cite{heusel2017gans}. The two best results are underlined; the best result is  bold.} 
\label{tab:similarity quantitative table}
\end{table}

\begin{figure*}[thpb]
    \centering
    \includegraphics[width=\linewidth]{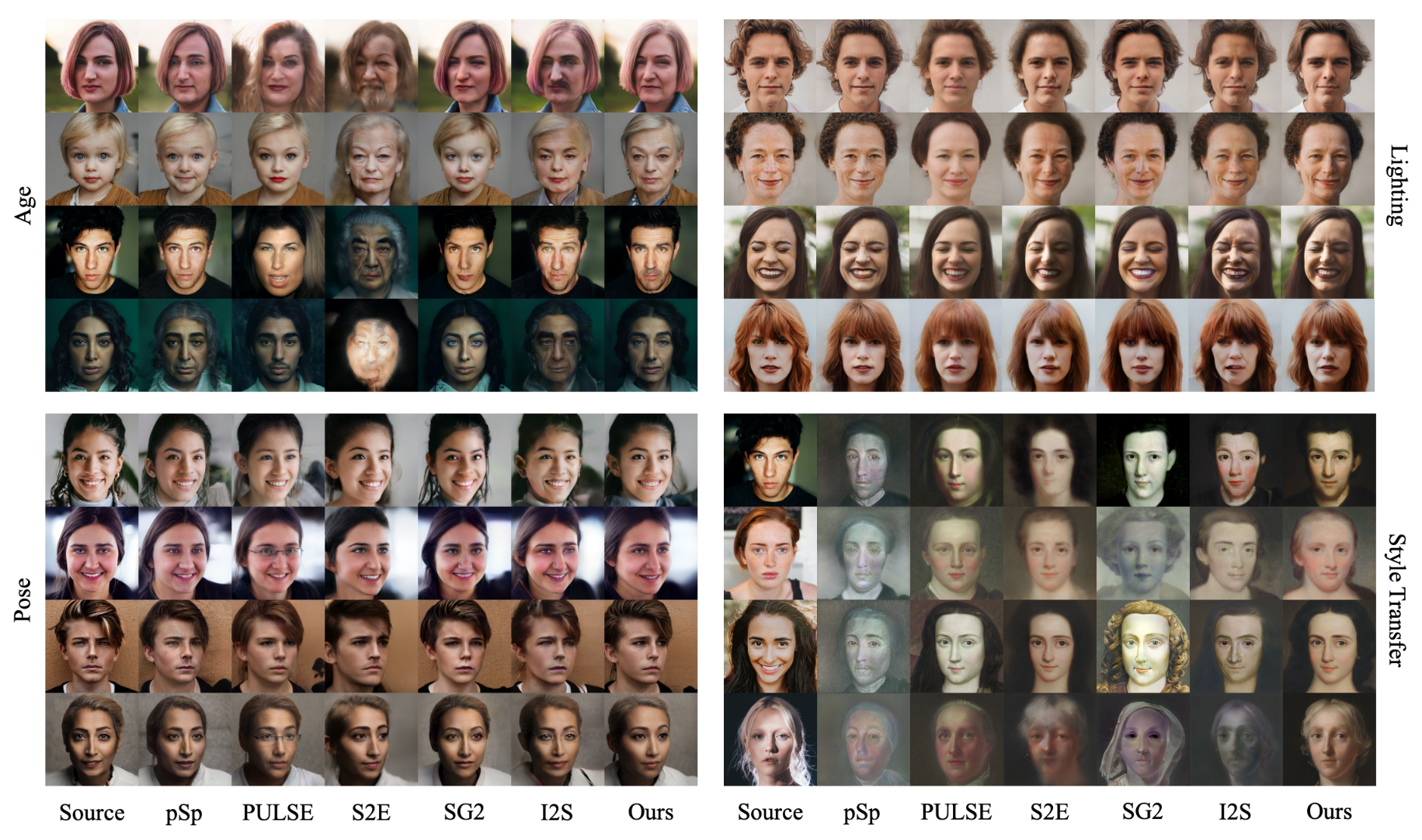}
   
     \caption{Four selected examples each for editing age, pose, lighting, and style transfer. Close inspection shows that II2S (Ours) consistently accomplishes the edit and also maintains the highest level of realism and also similarity to the source image after edits; this is consistent with a user study.}
    \label{fig:editing tasks}
\end{figure*}

\begin{figure}[thpb]
    \centering
    \includegraphics[width=\linewidth]{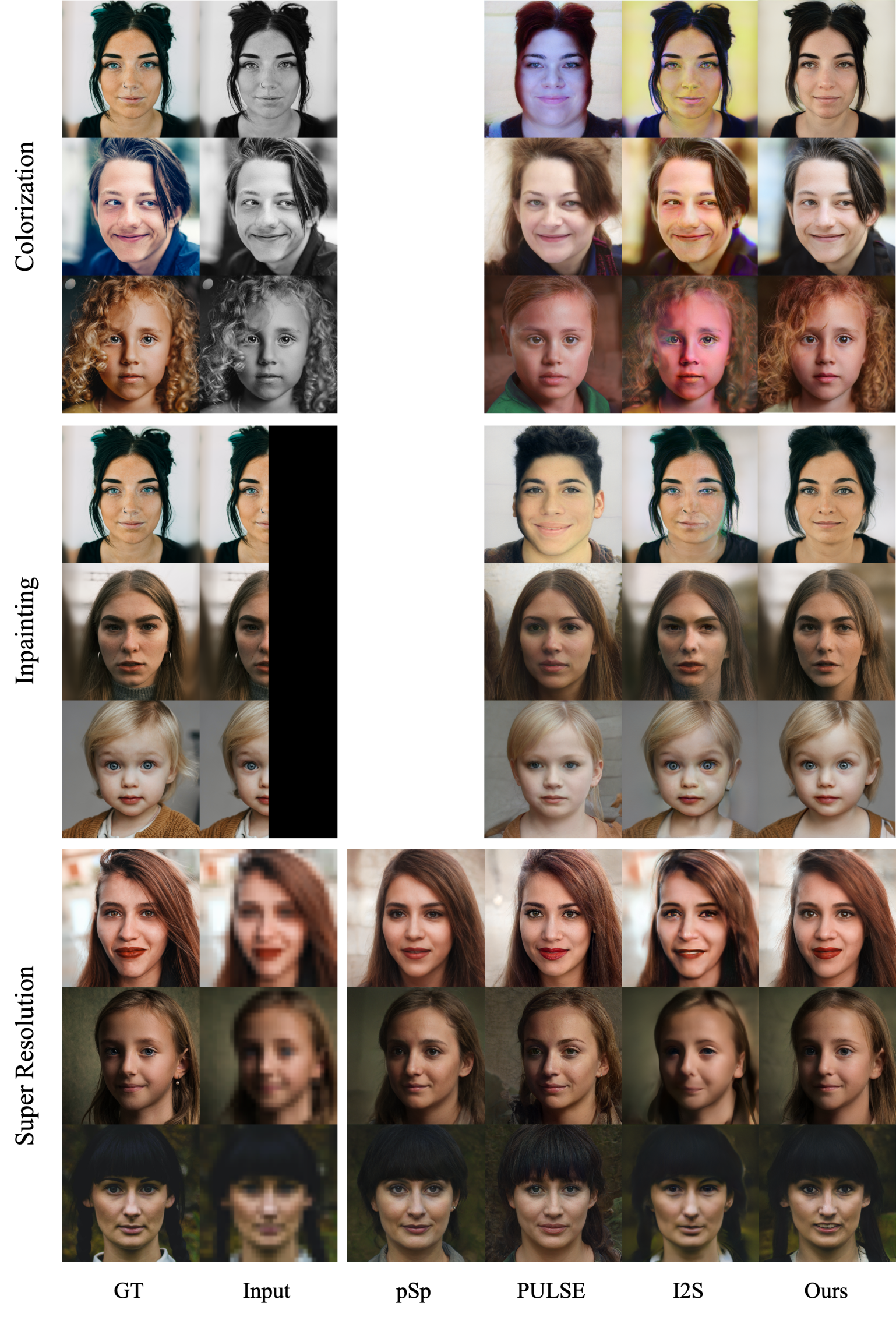}
   
     \caption{Examples of conditional embedding results. Three example of each task are shown from top to bottom: Image colorization, inpainting, and super resolution.  For pSp we only include their super-resolution results because they only included model weights for that task. Close inspection will show that II2S results satisfy the conditions with most realistically reconstructed images.}
    \label{fig:conditional embedding}
\end{figure}

\subsection{Editing Quality}
\label{sec:editing}

Unlike reconstruction, the editing quality of an embedding has not been studied, because competitive editing frameworks just became available very recently~\cite{shen2020interfacegan,harkonen2020ganspace,tewari2020stylerig,abdal2020styleflow}.
Note that these editing frameworks are orthogonal to our work, as they do not propose new embedding algorithms.
We choose StyleFlow~\cite{abdal2020styleflow} as our main evaluation method for editing as it generally leads to the highest quality edits.


We propose to test the editing quality using the following editing operations: pose, age, and lighting.
For the pose editing task, we set the target yaw of each image to 20$^\circ$. 
To edit age, we set the target age to 50 years old. 
In lighting change experiments, we set the target light direction to point towards the viewer's right. 
In Fig.~\ref{fig:editing tasks}, we visually compare editing results of different methods. Notice how our method keeps the identity of the original image and preserves the realism after the edits.
%

The user study results presented in Table~\ref{tab:user-studies} show that editing operations on latent codes derived by II2S maintain image quality across different editing tasks. Our editing results are preferred to the edits of other methods for all tasks. Somewhat surprisingly, SG2  is the second-best method for editing.
Based on our visual analysis of the results, we observe the following problem.
Some  embedding algorithms compute latent codes that cannot be edited well. 
After an edit, the new latent code may produce a new image that is very similar and did not fully accomplish the editing task. 
For example, when performing an edit to change the age, PULSE produces results that look noticeably younger than the target age.
We further study this phenomenon using the state-of-the-art classifier Microsoft Face API \cite{API} in supplementary materials.

\subsection{Conditional Embedding  Quality}

We consider four conditional embedding applications: image colorization, inpainting, super resolution, and style transfer.  
%
Let $I$ be an input ``condition'' image, and $G$ be a StyleGAN generator. Conditional embedding aims to locate an optimal latent code ${\mathbf{w}^{+}}^*$ in $W^+$ space so that the embedded image $G({\mathbf{w}^{+}}^*)$ i) faithfully captures the ``condition'' of $I$ and ii) is a realistic face image.
Accordingly, we define conditional embedding as:
\begin{equation}
     {\mathbf{w}^{+}}^*=\arg \min _{\mathbf{w^+}} L\left(I, f\left(G\left(\mathbf{w^+}\right)\right)\right)+\lambda \cdot R(\mathbf{w^+})
\end{equation}
where $f$ is a ``condition'' function that modifies an image to satisfy a pre-defined condition (\eg grayscale), $L$ is a loss function measuring the similarity between two images (\eg pixel-wise $L2$ loss, perceptual loss), $\lambda$ is a hyperparameter, and $R$ is the regularizer.
Note that the only difference between ordinary and conditional embedding is the incorporation of $f$.

For image colorization, the input image $I$ is a grayscale image and the condition function $f$ coverts a color image to grayscale. 
For inpainting, the input image $I$ is an incomplete image and the condition function $f$ is a {\it mask} function that erases the pixels in a given region. In our tests, the missing region is half the image.
For super resolution, the input image $I$ is a $32 \times 32$ low-resolution image and the condition function $f$ is a {\it downsampling} function.

The style transfer task is accomplished using a different approach. 
Let $G$ and $G'$ be two StyleGAN generaters, $G$ is trained on the FFHQ dataset and $G'$ is a variant of $G$ fine-tuned on the MetFace~\cite{Karras2020ada} dataset. 
Our style transfer is implemented by embedding the input image into $G$ and then evaluating the resulting latent code using $G'$.

The corresponding user study results for all four applications are presented in Table~\ref{tab:user-studies}.
Examples of visual results for the first three tasks are shown in Fig.~\ref{fig:conditional embedding} and results for style transfer are shown in Fig.~\ref{fig:editing tasks}. 
From the user study we can note that users ranked our image colorization results to be the most successful. Surprisingly, they are ranked even more realistic than the ground truth images. We attribute this to the fact that several ground truth images are stylized photographs. For the superresolution results, we asked users which image was closest to the ground truth. Our result is clearly preferred to all published competitors, I2S and PULSE. While our results are also better than pSp, they are not statistically significant. For inpainting we also significantly outperform all competitors. This is the only result where our method clearly looses to the ground truth. We ascribe this to the fact that we do inpainting for a very large missing region (half the image). This is a very challenging task. For pSp we could not complete the colorization and inpainting comparisons because pSp needs to pre-train a separate encoder for each task. Corresponding encoders were not provided by the authors. Similarly, we did not compare to SG2 and S2E on the first three tasks. This would have required us to reimplement conditional embedding in TensorFlow.



\subsection{User Studies}


We carried out a series of user studies using Amazon Mechanical Turk (MTurk).
For each task, images were generated using II2S and a competing method. 
Each pair of images was presented to workers on Amazon Mechanical Turk twice, once in the order (II2S, Source, Other), and once in the reverse order, so that preferences for the image based on its position in the survey would not be an issue.  Users were given a prompt to select the image that best accomplishes the task with the fewest artifacts. Table \ref{tab:user-studies} shows the percentage of responses that selected a {\jcf II2s over a competing} method, and a number {\jcf above} 50\% means that the II2S solution was selected more frequently.
Each survey received 120 responses, so any number below 40\% or above 60\% is statistically significant at 95\% confidence.  II2S outperforms competing methods on most tasks, with only a few exceptions. Reconstruction results are better with I2S, however I2S is effectively the same approach without regularization, and thus one would expect it to do better at reconstruction and poorer for editing tasks. The pSp approach for editing age was favored by more users, however the results were not statistically significant. For conditional GAN tasks, II2S is outperformed by Ground Truth, which is expected.   To summarize, we argue that our evaluation demonstrates that our proposed regularizer has a significant impact on downstream applications.


\begin{table}[thpb]	
\centering
\small


\def\theirs#1{\textit{#1}}
\def\ours#1{\textbf{#1}}

\begin{tabular}{lrrrrrr} 
	\toprule
	                & GT             & I2S     	      & SG2          & S2E          &  PUL          & pSp            \\ \midrule
	Reconstruction  & --             & \theirs{39.2}  & \ours{58.3}  & \ours{71.7}  &  \ours{83.3} 	& \ours{55.0}   \\
	Edit:Age 		& --             & \ours{80.0}	  & \ours{57.5}  & \ours{97.5}  &  \ours{64.2} 	& \theirs{47.08} \\
	Edit:Pose 		& --             & \ours{69.2}    & \ours{55.0}  & \ours{79.2}  &  \ours{69.2} 	& \ours{58.4}    \\
	Edit:Light   	& --             & \ours{79.2}	  & \ours{64.2}  & \ours{88.3}  &  \ours{76.7} 	& \ours{52.94}   \\
	Style Trans. 	& --             & \ours{77.5}    & \ours{84.2}  & \ours{76.7}  &  \ours{73.3} 	& \ours{90.83}   \\ 
	Colorization    & \ours{64.4}    & \ours{95.0}    & --           & --           &  \ours{79.4}  & --             \\
	Inpainting      & \theirs{28.3}  & \ours{65.8}    & --           & --           &  \ours{75.8}  & --             \\
    Super Res.		& \theirs{43.3}  & \ours{94.17}   & --           & --           &  \ours{65.8}  & \ours{52.5}    \\
	\bottomrule
\end{tabular}
\caption{A user study comparing II2S to the Ground Truth (GT), pSp, StyleGan2 (SG2), StyleGAN2 Encoder (S2E), Pulse (PUL) and I2S. 
Cases where II2S is preferred are in bold. Excluding ground-truth, there are only two competing methods that outperform II2S for any task.  }	
\label{tab:user-studies}
\end{table}

\subsection{Ablation Study}
The most important parameter in our method is the strength of the regularizer. Based on many experiments, we manually selected three settings of $\lambda$ to compare to: $0.01$, $0.005$, and $0.001$. These settings were selected so that a visual difference is noticeable between {\jcf images generated from} the embeddings. In Fig.~\ref{fig:recon} we show the visual difference between $\lambda=0.005$ and $\lambda=0.001$. In Tab.~\ref{tab:similarity quantitative table} we show how $\lambda$ affects reconstruction metrics. However, reconstruction metrics are not as informative as human perception. 
A user study to evaluate different choices of $\lambda$ is shown in Table~\ref{tab:ablation-user-studies}, and based on the average user response we picked $\lambda=0.005$ for all tests in the paper as a trade-off between reconstruction and editing quality.

\begin{table}[htbp]
\centering
\small

\begin{tabular}{lrrr} 
	\toprule
	             & .01 	 & .005   	& 0.001   \\ \midrule
	Reconstruction     & 28.8  & 49.6 	& \textbf{71.7}  \\
	Edit:Age 	 & 56.7  & \textbf{60.0}    & 33.3   \\
	Edit:Pose 	 & 47.9  & 48.3 	&\textbf{ 53.8}   \\
	Inpainting   & 48.8  &\textbf{ 53.8} 	& 47.5 \\
    SuperResolution	 & \textbf{61.3 } & 53.3 	& 35.4   \\
    \midrule
    \textit{Average} & 53.6 &\textbf{ 53.9} & 42.5 \\
\end{tabular}

	\caption{User Study of $\lambda$. 	Each value is the percentage of times an image was selected as the best when compared to an image form \textit{either} of the other two columns; higher numbers are better.} 
	\label{tab:ablation-user-studies}
  
\end{table}	


\subsection{Histogram of the Embeddings}
\label{sec:histogram}

\begin{figure}[thpb]
    \centering
    \includegraphics[width=\linewidth]{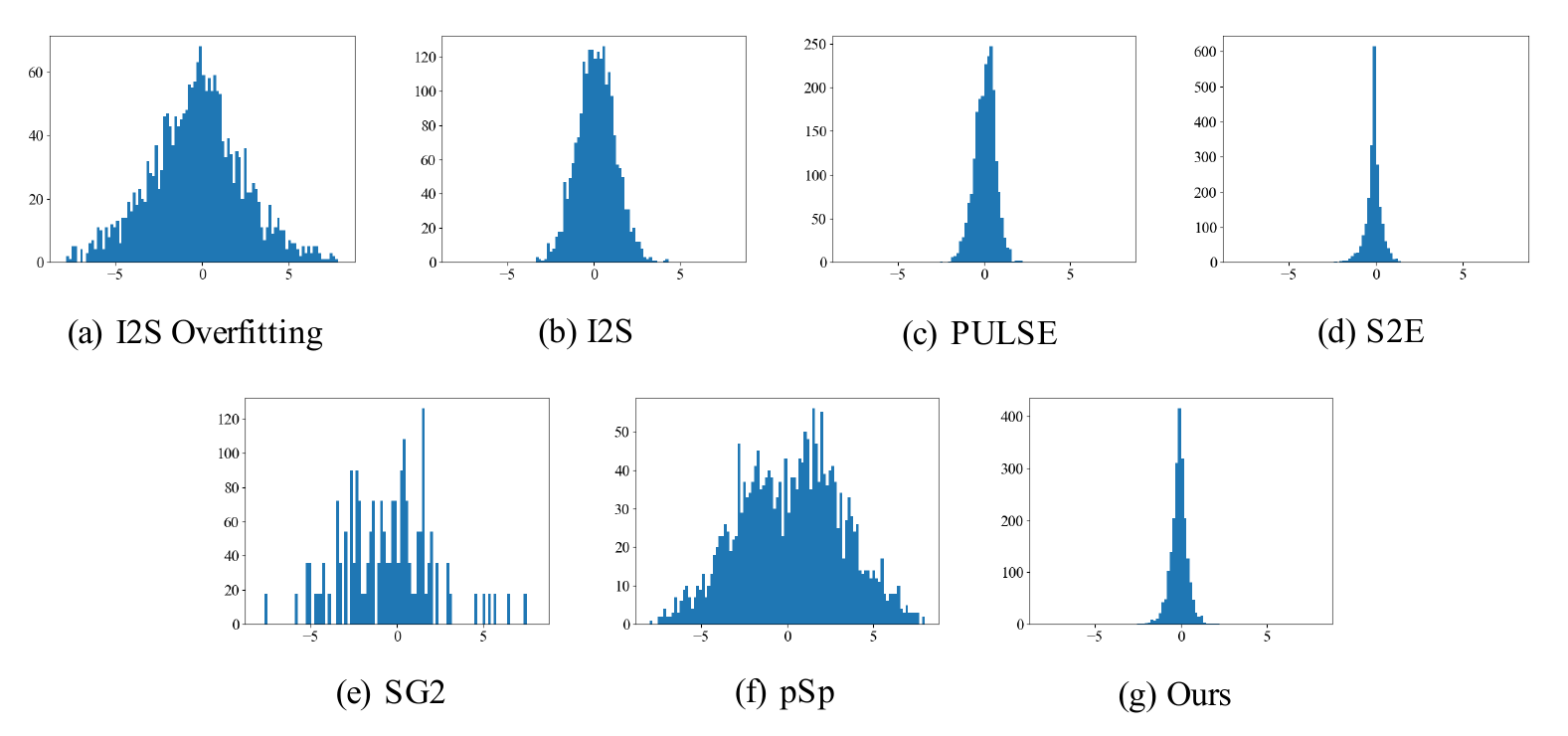}
    \caption{Histograms showing the distribution of latents {\jcf in an arbitrary (the twentieth)} dimension of the \Pnormplus{} Space.} 
    \label{fig:histo1}
\end{figure}

We visually analyze the histograms in the \Pnormplus{} space to obtain additional insights about the different embedding algorithms.
To compute the histograms we embed the 120 images of our dataset using the different algorithms. Then we take the $20^{th}$ dimension of each latent code in \Pnormplus{}{}. For all methods except SG2, there are 18 such latent codes per embedded image. For SG2, we repeat the same value 18 times.
In Fig.~\ref{fig:histo1} we show the histograms.
For the first histogram, denoted by \textit{I2S Overfitting}, we run I2S with a high learning rate. As we know that this setting leads to overfitting, we can observe that overfitting correlates with a histogram that is wider, has higher variance, and more outliers.
The histograms of the encoders, pSp and S2E, have the lowest variance, indicating that the embedded latent codes are closer to the origin of \Pnormplus{}{}. It is our conjecture that narrow histograms correlate with underfitting. The histogram for SG2 looks very strange. This might be due to the fact that SG2 has fewer unique samples. Still, the variance is significantly higher than we would have expected.

\section{Limitations, Future Work, and Conclusion}

Our work has some limitations that we leave to future work. 
One limitation is that we did not investigate correlations between the (18) different components of an extended latent space. From sampling new latent codes in $Z^{+}$ space, we know that sampling the 18 components independently leads to poor results.
However, it is unclear if their similarity should be enforced by an additional regularizer, as we could not really observe a consistent improvement when experimenting with different regularizers proposed in the literature (e.g. hierarchical optimization or an $L2$ similarity term). In the future, we would like to revisit this topic.

We introduced the \Pnorm{} space as a tool to facilitate improved StyleGAN embedding and to analyze different embedding algorithms. The switch to \Pnorm{} space can significantly help follow-up work. We proposed a new regularizer for StyleGAN embedding that provides the best trade-off between reconstruction quality and editing quality. 
Finally, we conducted an extensive evaluation, highlighting the strength and weaknesses of previous work. The evaluation demonstrates that our results clearly outperform all competing methods for downstream applications.

%
%
%
%

\clearpage
\newpage
\bibliographystyle{ACM-Reference-Format}
\bibliography{sample-bibliography}

\appendix

\section{Metrics}

\subsection{Comparison of Editing Quality}

We choose StyleFlow~\cite{abdal2020styleflow} as our main evaluation method for editing as it generally leads to the highest quality edits. In addition, we compare against another editing approach, \- GANSpace~\cite{harkonen2020ganspace} in this supplemental material.

We propose to test the editing quality using the following editing operations: pose, age, lighting, gender and eyeglasses.
For the pose editing task, we set the target yaw of each image to 20$^\circ$. 
To edit age, we set the target age to 50 years old. 
In lighting change experiments, we set the target light direction to point towards the viewer's right. For the gender transfer task, we swap the gender of the input image. In eyeglasses experiments, we add eyeglasses to each image. Note that the only conditional edit is the gender swap. Here the goal of the edit depends on the input image. For other edits, there is a fixed goal, such as adding eyeglasses. If the person had eyeglasses before, the image is not supposed to change. Similarly, for the pose, there is a fixed target pose. Some images are expected to change more than others under these edits. Images that are already in the target pose are expected not to change.

\begin{figure}[th]
    \centering
    \includegraphics[width=\linewidth]{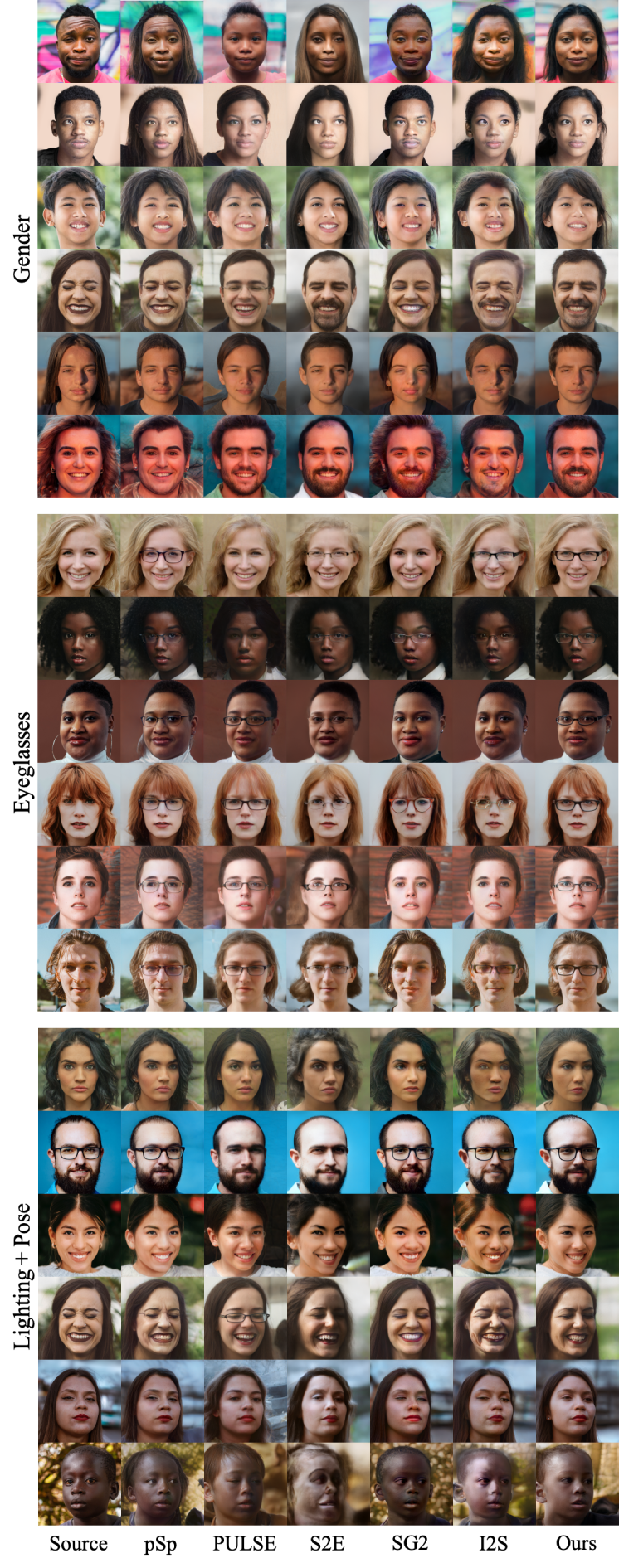}

      \caption{Examples of editing gender, eyeglasses and lighting + pose. The results of edits using the proposed \Pnormplus{} regularizer are shown on the right. Compared to other methods, our results tend to preserve non-edited attributes/identity of the input, generate realistic images, and also accomplish the editing task. }
    
    \label{fig:Supp_editing}
\end{figure}

We already show examples of pose, age, and lighting in the main paper.
In Fig.~\ref{fig:Supp_editing}, we visually compare the remaining editing results (gender, eyeglasses and lighting+pose) of different methods. Notice how our method keeps the identity of the original image and preserves the realism after the edits.
Quantitative results for various edits are presented in Table~\ref{tab:editing table}. To create the table, attributes were measured using Microsoft's face API~\cite{API}. Lighting is measured by a different network~\cite{DPR}. Overall, our method has the best results. pSp is very slightly better than ours in two edits, but clearly worse in two others.

\begin{table}[thpb]
\centering
\footnotesize
\begin{tabular}{cccccc} \toprule

\multirow{2}{*}{Method}  & \multicolumn{1}{c}{Age}  & \multicolumn{1}{c}{Pose}  & \multicolumn{1}{c}{Lighting} & \multicolumn{1}{c}{Gender} & \multicolumn{1}{c}{Eyeglasses} \\\cmidrule{2-6}

 & \multicolumn{1}{c}{50}  & \multicolumn{1}{c}{20}  & \multicolumn{1}{c}{R$\rightarrow$L} & \multicolumn{1}{c}{Swap} & \multicolumn{1}{c}{Add}  \\ \midrule
                       
pSp~\cite{richardson2020encoding} &38.25  &\underline{18.2} &0.87 &0.78  &\underline{\textbf{1.00}} \\

PUL~\cite{menon2020pulse}  &33.8	&\underline{\textbf{18.3}}	&0.88	&0.71	&0.76 \\

S2E~\cite{stylegan2encoder}   &60.1	&25.7	&\underline{0.91}	&\underline{0.93}	&0.83\\

SG2~\cite{styleganv2-2020} &29.4	&13.2	&0.85	&0.33	&0.38\\


I2S~\cite{image2stylegan2019} &\underline{53.4}	&18.1	&0.90	&0.85 &0.71\\

Ours &\underline{\textbf{50.2}}	&23.5	&\underline{\textbf{0.94}}	&\underline{\textbf{0.99}} &\underline{0.98}\\

\bottomrule
\end{tabular}


\caption{
Quantitative comparison of editing quality.} 
\label{tab:editing table}
\end{table}

\begin{figure}[th]
    \centering
    \includegraphics[width=\linewidth]{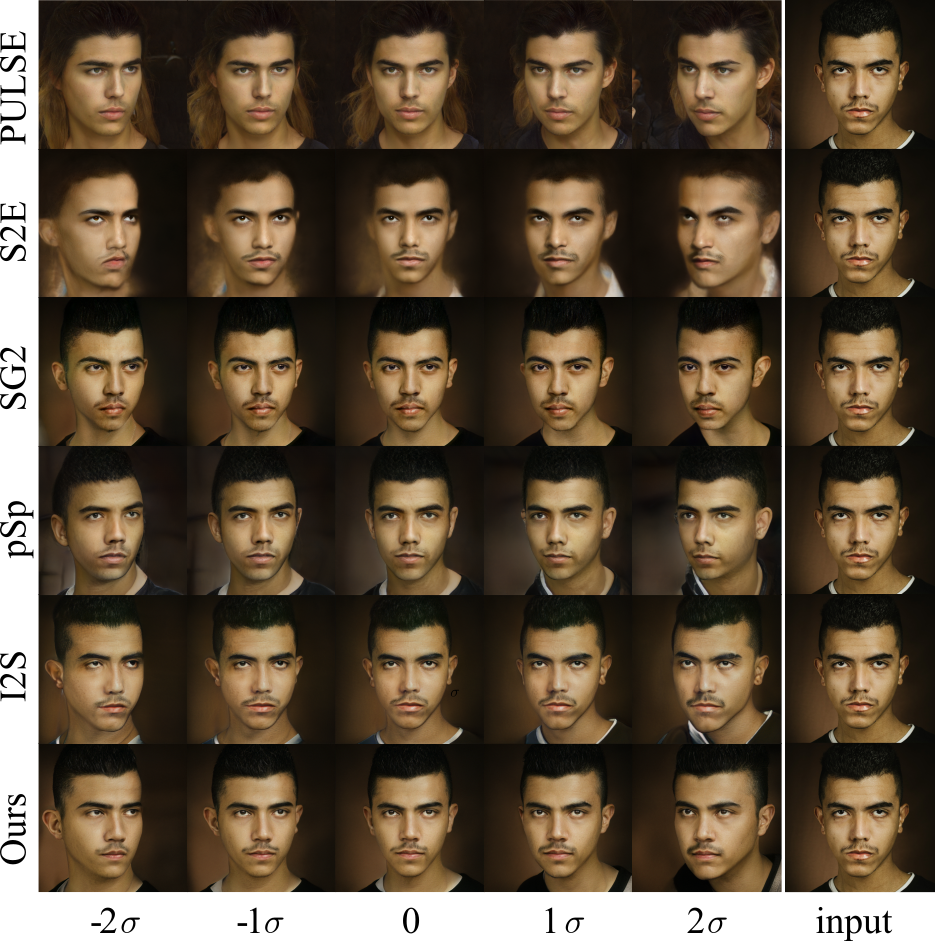}

      \caption{Examples of pose editing using GANspace~\cite{harkonen2020ganspace}. Our approach does a better of reconstructing the input than many methods (notice the facial hair) and it also accomplished the editing task. Compared to I2S, which also does well at reconstructing the image near the input pose, our approach maintains realism at the extreme poses. For example, the nose position in 3D, and the shape of the iris are preserved well in ours. }
    
    \label{fig:Supp_ganspace}
\end{figure}

In addition to using StyleFlow~\cite{abdal2020styleflow} as the evaluation method of editing quality, for the pose change experiments, we use the right-left pose edit from GANSpace~\cite{harkonen2020ganspace}. Specifically, for a latent code in the GANSpace coordinate system, we can set the coordinate corresponding to pose to five different values: -2$\sigma$, -$\sigma$, 0, $\sigma$, and +2$\sigma$, where $\sigma$ is the eigenvalue for the direction. Fig.~\ref{fig:Supp_ganspace} shows the qualitative results of pose change edits. Quantitative results are presented in Table~\ref{tab:POSE table}. We observe that the strength of an edit depends on the initial latent code and the embedding method. This poses a challenge for the evaluation of different embedding methods using GANSpace.

\begin{table}[thpb]
\centering
\small
\begin{tabular}{lrrrrr} \toprule

\multirow{1}{*}{Method}  & \multicolumn{1}{c}{-2$\sigma$}  & \multicolumn{1}{c}{-1$\sigma$}  & \multicolumn{1}{c}{0} &  \multicolumn{1}{c}{1$\sigma$}& \multicolumn{1}{c}{2$\sigma$}\\ \midrule
                       
pSp
&19.40	&10.48	&-0.58	&-11.58	&-20.63\\
PUL
&20.12	&10.16	&-0.31	&-9.66	&-17.99\\
S2E
&\underline{25.75}	&11.66	&-1.34	&-13.52	&\underline{\textbf{-26.35}}\\
SG2
&12.49	&4.84	&-2.04	&-8.36	&-14.75\\

I2S 
&18.42 	&8.29	&-0.81	&-9.27	&-18.49    \\

Ours
&\underline{\textbf{26.23}}	&13.16	&-0.01	&-12.07	&\underline{-23.42}
\\
\bottomrule
\end{tabular}
\caption{Pose metric for testing the editing quality using GANspace~\cite{harkonen2020ganspace}. We mark the highest ranges. These numbers by themselves are not conclusive with regard to quality, but we can observe that our method and pSp react stronger to this GANSpace edit than other methods. We can also observe that the dependence of the strength of an edit on the initial embedding is quite strong.}
\label{tab:POSE table}
\end{table}

\subsection{Linear Interpolation}

\begin{figure*}[th]
    \centering
    \includegraphics[width=\linewidth]{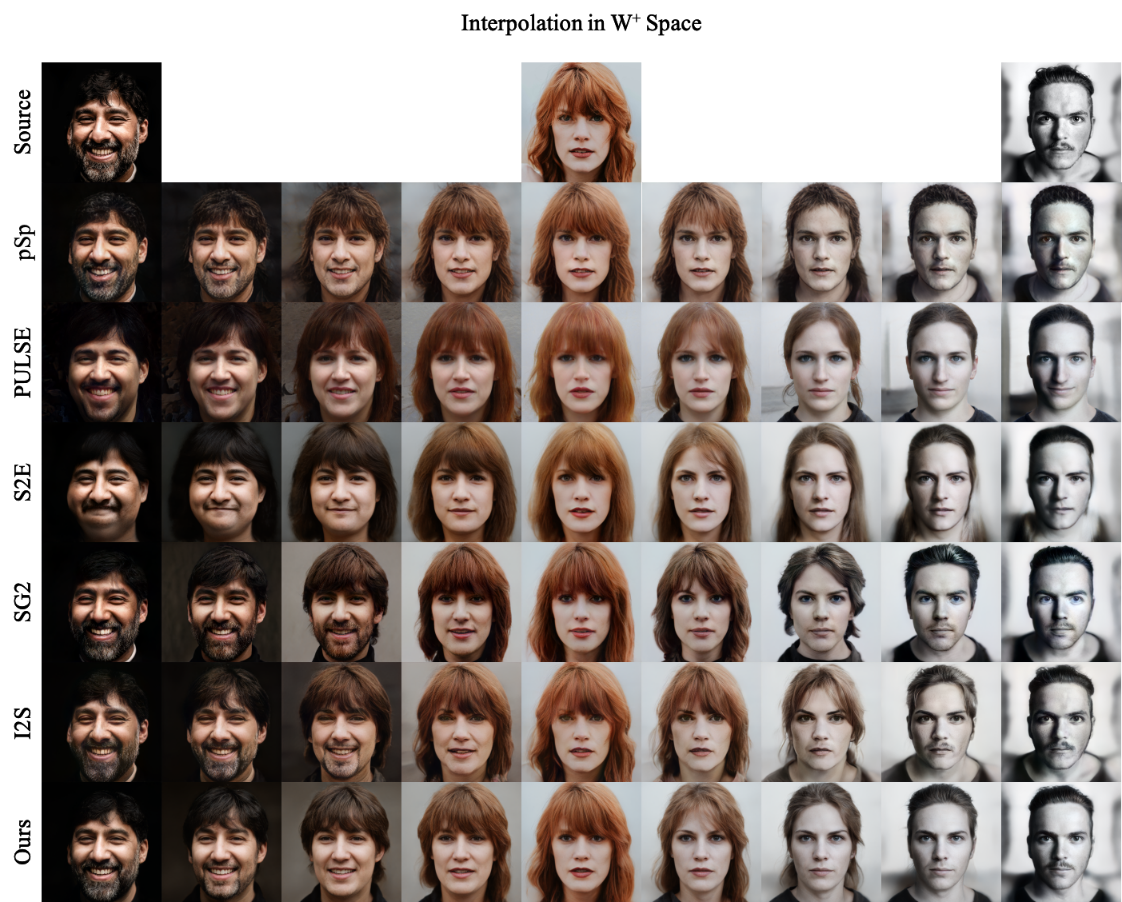}

      \caption{Examples of Interpolation in \Wplus{} space. This shows two linear interpolations, interpolating from column one to column five, and then from column five to column ten. The last row uses our embeddings - notice that the initial reconstruction quality (columns one, five, ten) is high, and that in-between images remain plausible. }
    
    \label{fig:Supp_Interpolation_W}
\end{figure*}

\begin{figure*}[th]
    \centering
    \includegraphics[width=\linewidth]{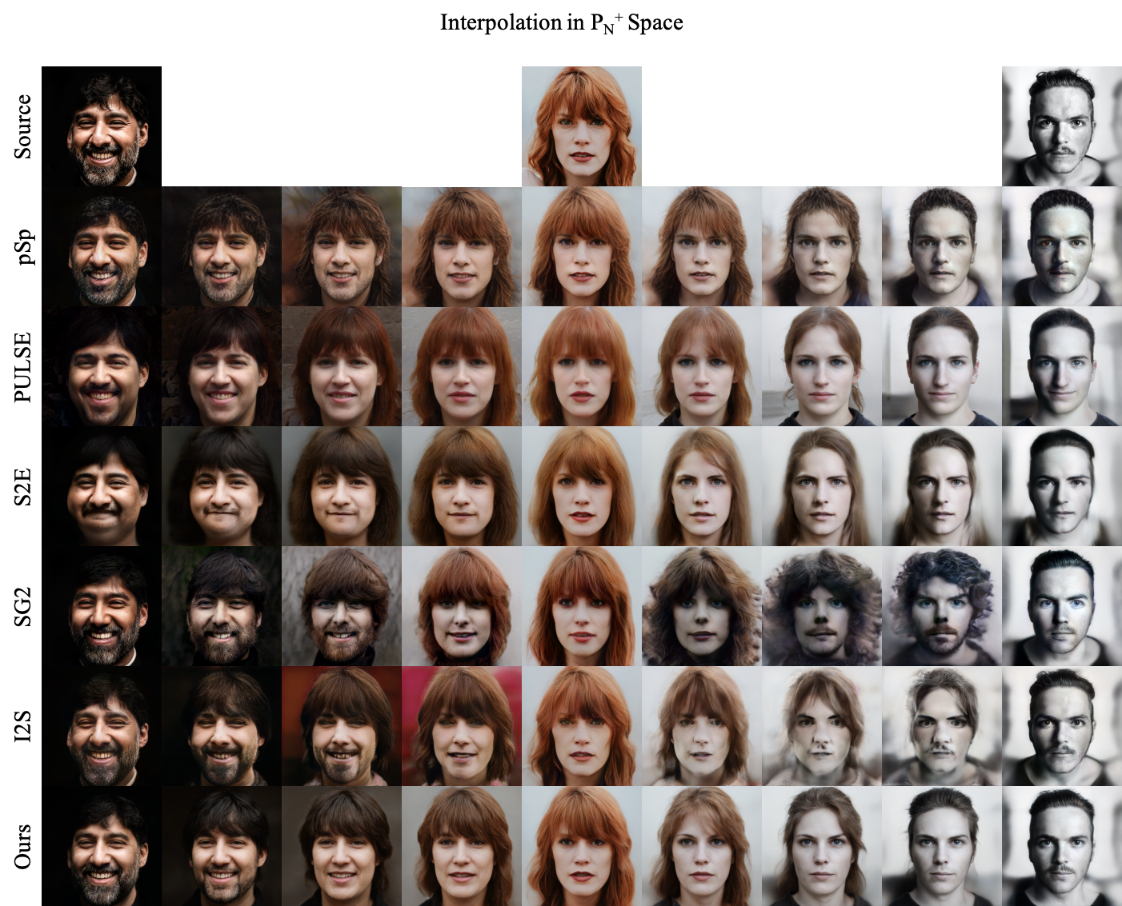}

      \caption{Examples of Interpolation in \Pnormplus{} space. This shows two linear interpolations, one from the top-left image to the center, and the other interpolation from top-center to the top-right image.  We do not recommend actually performing edits in this space -- \Pnormplus{}~space is useful for \textit{finding} latents in dense regions of the mapping network, however the actual edits produce more intuitive results if the are carried out in a disentangled space such as \Wplus{}.}
    
    \label{fig:Supp_Interpolation_P_N}
\end{figure*}

Linear interpolation of latent codes can generate a morph between two embedded images. It is also a powerful tool to determine whether an embedding is semantically meaningful.
In other words, we can tell if an embedding is good if the interpolated images between it and a randomly sampled image are of high quality.
To this end, we test the interpolation results in the $W^+$ space.
Specifically, we compute the FIDs between the 120 images of our dataset and 2,500 interpolated images obtained by i) sampling 500 image pairs from the dataset and ii) interpolating 5 images for each pair at proportions 0.1, 0.25, 0.5, 0.75 and 0.9.
Table~\ref{tab:interpolation_table} shows the results on various methods. 
Qualitative results are shown in Fig.~\ref{fig:Supp_Interpolation_W}. 
We propose the use of \Pnormplus{} as a regularizer, however it is a nonlinear mapping from \Wplus{} space,  which can negatively impact entanglement.  
However, we show the results of linear interpolation in Fig.~\ref{fig:Supp_Interpolation_P_N}, While we have the best quantitative results together with I2S one has to consider that FID is only calculated on a smaller set of images (2500 instead of 50000). While several other authors also report FID on smaller sets of images for editing and manipulation tasks, the metric is not entirely conclusive. For the qualitative results, it is important to mention that our method has the best reconstruction quality.



\begin{table}[thpb]
\centering
\footnotesize
\begin{tabular}{ccccc} \toprule

\multirow{1}{*}{Method} & \multicolumn{1}{c}{Style Mixing} & \multicolumn{1}{c}{Interp. in $W^{+}$} &\multicolumn{1}{c}{Interp. in \Pnormplus{}}  \\\midrule

pSp     &65.45  &57.93  &58.74	\\

PUL 	&65.05	&63.73    &63.39 \\

S2E 	&66.22	&	67.84 &69.91\\

SG2 	&58.48	&49.40 &\underline{54.19}\\

I2S &\underline{57.36}	&\underline{\textbf{43.03}}   	&55.25    \\

Ours	&\underline{\textbf{55.69}}	&\underline{48.61}	&\underline{\textbf{49.15}}\\

\bottomrule
\end{tabular}


\caption{
FID of style mixing and interpolation in different spaces.} 
\label{tab:interpolation_table}
\end{table}

\begin{figure}[thpb]
    \centering
    \includegraphics[width=\linewidth]{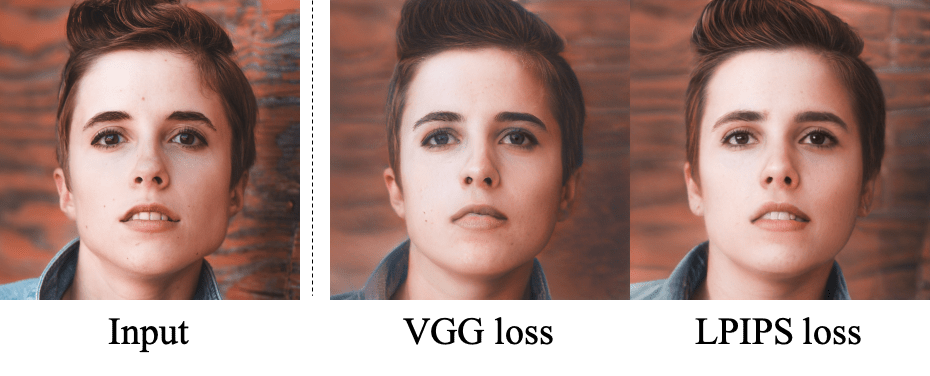}
    \caption{LPIPS perceptual loss vs. VGG perceptual loss. The original I2S used VGG-loss, but LPIPS encourages better reconstruction of details. }
    \label{fig:Supp_diffetent_loss}
\end{figure}

\begin{figure}[thpb]
    \centering
    \includegraphics[width=\linewidth]{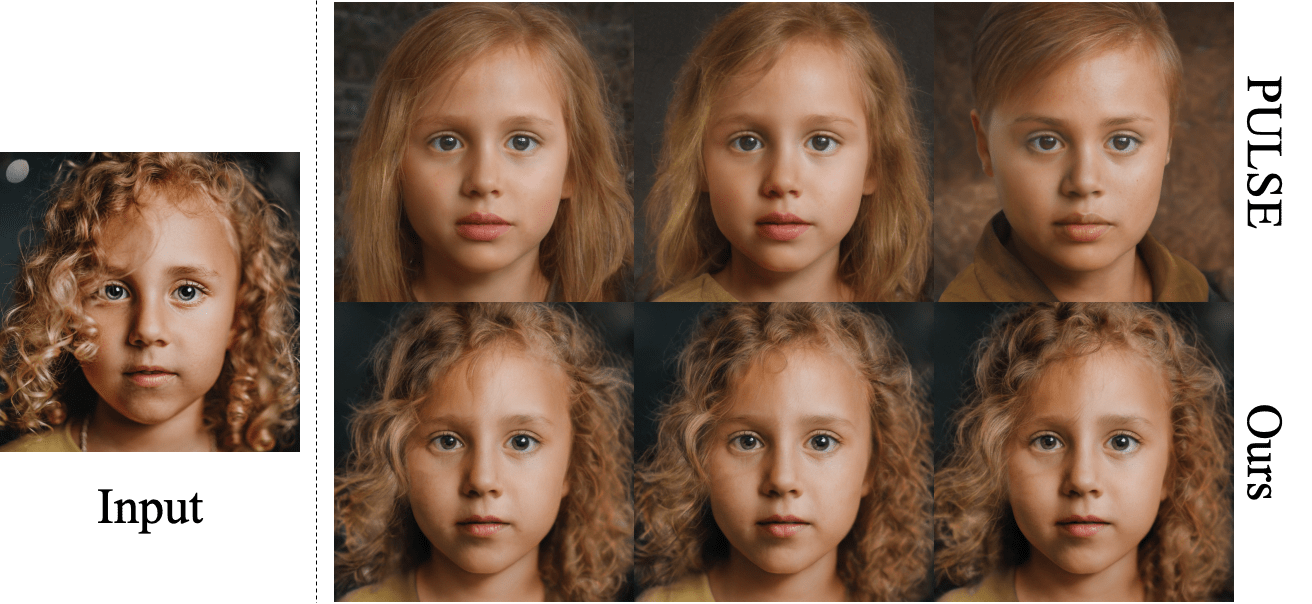}
    \caption{Comparing the stability of the algorithms using repeated embeddings. Each run of PULSE will provide different results, while our results are relatively stable. This is a visual demonstration of the problem that the PULSE hypersphere constraint creates. Our approach has no such constraint.}
    \label{fig:Supp_stable}
\end{figure}

\begin{figure}[thpb]
    \centering
    \includegraphics[width=\linewidth]{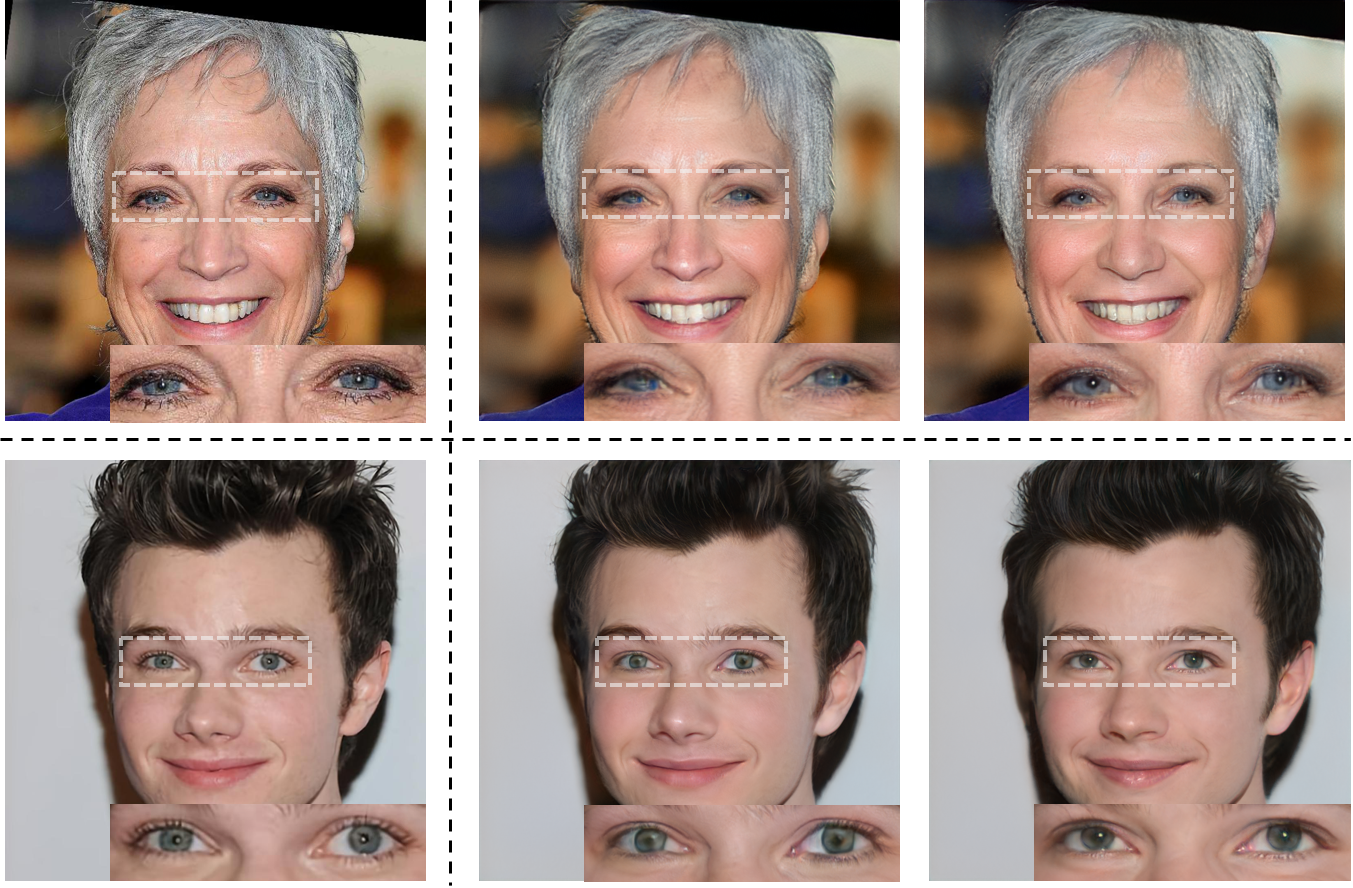}
    \caption{Eye correction obtained by applying the regularizer in the \Pnormplus{} space on $W^{+}$ embedding. Left: original image, middle: $W^{+}$ embedding, right: $W^{+}$ embedding with \Pnormplus{} regularizer. Without our regularization, the generator is not able to preserve the shapes of the pupil and iris. }
    \label{fig:eye_correct}
\end{figure}

\subsection{Style Mixing}

As two sides of the same coin, the results of style mixing and structural edits (\eg pose editing) both reflect how good a latent disentangles the style and contents at different layers of the StyleGAN generator.
Thus, it is also worthwhile to check whether an embedding enables high-quality style mixing. We take the variables from the first 7 layers from one image and from the remaining 11 layers from another image and test the FID of the results with the original images.

\begin{figure*}[th]
    \centering
    \includegraphics[width=0.75\linewidth]{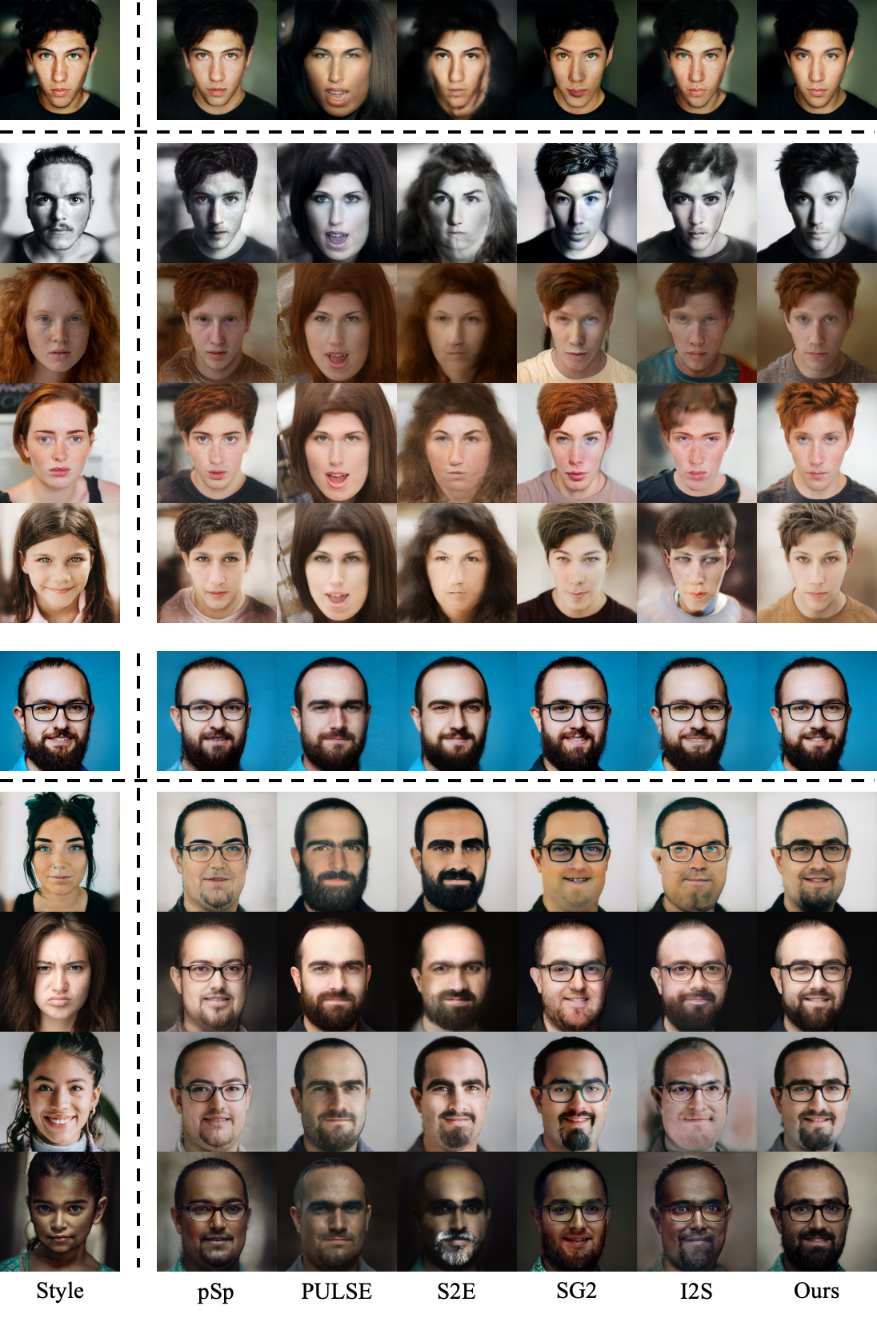}

      \caption{Examples of style mixing.}
    
    \label{fig:Supp_style_mixing}
\end{figure*}

Table~\ref{tab:interpolation_table} shows the result of style mixing on various methods. Qualitative results are shown in Fig.~\ref{fig:Supp_style_mixing}. Again, even though our method has the best results, FID is calculated for a smaller set of 500 images. Also, for the qualitative results, it's important to consider that our method has the best reconstruction quality.

\section{Ablation Study}

We conducted two additional ablation studies for the supplementary materials. The first ablation study compares the results when embedding in $Z^+$ space rather than $W^+$ space. While other authors advocate for embedding in $Z^+$ space, we obtain clearly worse results in our implementation. Specifically, we observe artifacts for several of the embedded images. A quantitative evaluation in shown in Table~\ref{tab:different_space}.

Another ablation study investigates the result of the different downsampling operation and the effect of choosing LPIPS as regularizer over VGG. The different downsampling operation has a very minor effect on the results, but exchanging VGG for LPIPS leads to significant and important differences. We show quantitative results in Table~\ref{tab:different_loss} and qualitative results in Fig.~\ref{fig:Supp_diffetent_loss}.

\begin{table}[thpb]
\centering
\footnotesize
\begin{tabular}{cccccccc} \toprule

\multirow{1}{*}{Space}  & \multicolumn{1}{c}{SSIM}  & \multicolumn{1}{c}{RMSE}  & \multicolumn{1}{c}{PSNR} & \multicolumn{1}{c}{VGG} & \multicolumn{1}{c}{LPIPS}  & \multicolumn{1}{c}{FID}  \\ \midrule

$Z^{+}$ &0.79 &0.10 &19.99 &0.99 &0.31 &63.57\\
$W^{+}$ &\textbf{0.83}	&\textbf{0.07}	&\textbf{23.00}	&\textbf{0.76}	&\textbf{0.20} &\textbf{43.99}\\

\bottomrule
\end{tabular}


\caption{
Ablation study of optimization in different spaces. The hyperparameter $\lambda$ of the regularizer  is set to 0.001.} 
\label{tab:different_space}
\end{table}

\begin{table}[thpb]
\centering
\footnotesize
\begin{tabular}{ccccccc} \toprule

\multirow{1}{*}{Method}  & \multicolumn{1}{c}{SSIM}  & \multicolumn{1}{c}{RMSE}  & \multicolumn{1}{c}{PSNR} & \multicolumn{1}{c}{VGG} & \multicolumn{1}{c}{LPIPS}   \\ \midrule
                       
Bilinear down. &0.82 &0.08 &22.95 &0.77 &0.21 \\
VGG percep. &0.82 &0.08 &22.45 &0.70 &0.24 \\
Ours &\textbf{0.83}	&\textbf{0.07}	&\textbf{23.00}	&\textbf{0.76}	&\textbf{0.20}\\
\bottomrule
\end{tabular}


\caption{
Ablation study of perceptual loss and down sampling method. The hyperparameter $\lambda$ of the regularizer  is set to 0.001.} 
\label{tab:different_loss}
\end{table}

\section{Looking at the ``Eyes'' and ``Mouth''}
\begin{figure*}[thpb]
    \centering
    \includegraphics[width=\linewidth]{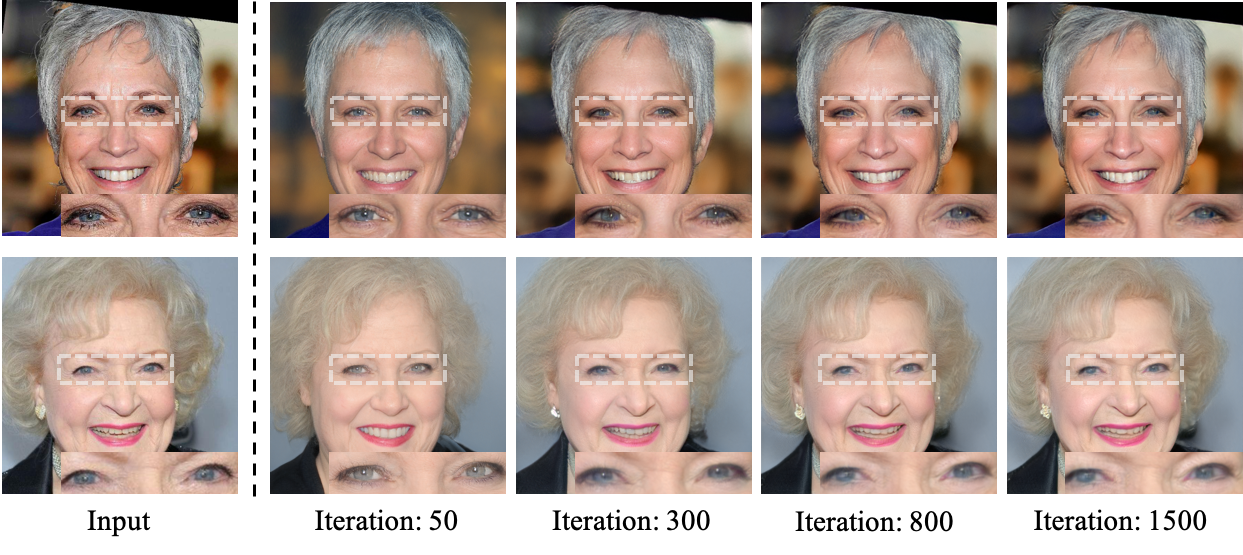}
    \caption{Progression of embedding using I2S. {\bf Notice that the eyes eventually lose their realism with 800 iterations.} The original image is shown to the left.}
    \label{fig:eyes}
\end{figure*}

Several existing evaluations in previous work may overemphasize reconstruction quality in the hyperparameter settings, when the goal it editing. For example, we could observe that Zhu et al.~\cite{zhu2020indomain} and Tewari et al.~\cite{tewari2020pie} use suboptimal I2S~\cite{image2stylegan2019} hyperparameter settings for some comparisons.
One simple test is to check the visual realism of the eyes and teeth during the embedding to observe the transition from reasonable editing quality to overfitting and poor editing quality.
In Fig.~\ref{fig:eyes} we show a progression of the embedding process with closeup insets of the eyes of the original I2S algorithm. This simple test is of high practical importance.
Our proposed regularizer is very important, because it is too difficult and impractical to set the number of iterations for each image separately for I2S. While it is too time-consuming to do this for our complete dataset, we believe that I2S results would improve using per image fine-tuning.
See Fig.~\ref{fig:eye_correct} for a visualization how the regularizer can prevent overfitting. This makes a big difference for eyes and teeth.

\section{Visual Explanation of the Pulse Regularizer Problem}

To better illustrate the problem of the Pulse regularizer, we perform an experiment of repeatedly embedding the same image. For most methods, embedding the same image multiple times leads to almost the same result. However, the Pulse regularizer uses a projection onto a subset of the latent space and a random initialization. This Pulse subset has high quality, but overall the restriction is unnecessarily restrictive. Also, the initial starting point and the location of the initial projection has a big influence on the final result. We believe Pulse is the only method embedding method that gives widely different results for the same input. Also, if the random initialization of Pulse is unlucky, the result of the embedding will be very different from the input image. See Fig.~\ref{fig:Supp_stable} for an example.

\begin{figure*}[thpb]
    \centering
    \includegraphics[width=\linewidth]{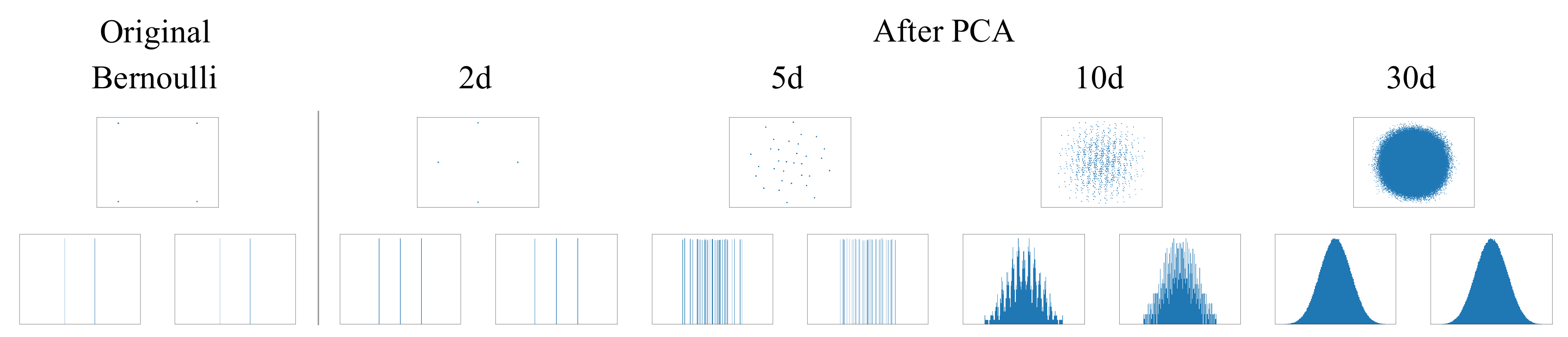}
    \caption{Marginal and pair-wise distributions of Bernoulli distribution in different dimensions before and after PCA. Only show the first two dimensions after PCA.}
    \label{fig:bernoulli}
\end{figure*}

\end{document}